\documentclass[10pt,journal,compsoc]{IEEEtran}

\newcommand{\ie}{\textit{i}.\textit{e}.}
\newcommand{\eg}{\textit{e}.\textit{g}.}

\usepackage{hyperref}       % hyperlinks
\hypersetup{
colorlinks=true,
linkcolor=red,
anchorcolor=green,
citecolor=green}

\usepackage{bbm}
\usepackage{pifont}
\usepackage{xcolor}

\definecolor{mygray}{gray}{0.4}

\ifCLASSOPTIONcompsoc
  \usepackage[nocompress]{cite}
 \usepackage{amsmath,amsfonts}
  \usepackage{algorithmic}
  \usepackage{algorithm}
  \usepackage{array}
  \usepackage{textcomp}
  \usepackage{stfloats}
  \usepackage{url}
  \usepackage{verbatim}
  \usepackage{graphicx}
  \usepackage{cite}
  \usepackage{multirow}
  \usepackage{amsthm}
  \usepackage{booktabs}
  \usepackage{colortbl}
  \usepackage{caption}
  \usepackage{multirow}
\usepackage{booktabs}
  \usepackage{subcaption}
  
\else
  \usepackage{cite}
\fi

\ifCLASSINFOpdf
\else
\fi

\usepackage{amsmath}
\usepackage{array}
\usepackage{pifont}

\usepackage{fixltx2e}
\usepackage{stfloats}
\usepackage{tikz}
\newcommand*{\circled}[1]{\lower.7ex\hbox{\tikz\draw (0pt, 0pt)%
    circle (.4em) node {\makebox[1em][c]{\small #1}};}}

\usepackage{url}

\begin{document}

\title{MotionVerse: A Unified Multimodal Framework for Motion Comprehension, Generation and Editing}

\author{Ruibing~Hou, %~\IEEEmembership{Member,~IEEE},
        Mingshuang~Luo, %~\IEEEmembership{Student Member,~IEEE},
        Hongyu~Pan,
        Hong~Chang*, %~\IEEEmembership{Member,~IEEE}, 
        and~Shiguang~Shan~\IEEEmembership{Fellow,~IEEE}
\IEEEcompsocitemizethanks{\IEEEcompsocthanksitem R. Hou and M. Luo are with Key Laboratory of Intelligent Information Processing, Institute of Computing Technology (ICT), Chinese Academy of Sciences (CAS), Beijing, 100190, China. 
 % \IEEEcompsocthanksitem  H. Pan is in Horizon Robotics.
 \IEEEcompsocthanksitem  H. Chang, and S. Shan are with Key Laboratory of Intelligent Information Processing, Institute of Computing Technology (ICT), Chinese Academy of Sciences (CAS), Beijing, 100190, China, and University of the Chinese Academy of Sciences, Beijing 100049, China.
\IEEEcompsocthanksitem *:Corresponding authors.
 \protect\\
E-mail: \{houruibing, changhong, sgshan\}@ict.ac.cn, mingshuang.luo@vipl.ict.ac.cn, hongyu.phy@gmail.com}% <-this % stops an unwanted space

%\thanks{Manuscript received April 19, 2005; revised August 26, 2015.}
}

% The paper headers
\markboth{Journal of \LaTeX\ Class Files,~Vol.~14, No.~8, August~2015}%
{Shell \MakeLowercase{\textit{et al.}}: Bare Demo of IEEEtran.cls for Computer Society Journals}

\IEEEtitleabstractindextext{%
\begin{abstract}
This paper proposes MotionVerse, a unified framework that harnesses the capabilities of Large Language Models (LLMs) to comprehend, generate, and edit human motion in both single-person and multi-person scenarios. To efficiently represent motion data, we employ a motion tokenizer with residual quantization, which converts continuous motion sequences into multi-stream discrete tokens. 
Furthermore, we introduce a \textit{Delay Parallel} Modeling strategy, which temporally staggers the encoding of residual token streams. This design enables LLMs to effectively capture inter-stream dependencies while maintaining computational efficiency comparable to single-stream modeling. Moreover, to alleviate modality interference between motion and language, we design a \textit{dual-tower architecture} with modality-specific parameters, ensuring stable integration of motion information for both comprehension and generation tasks. Comprehensive ablation studies demonstrate the effectiveness of each component in MotionVerse, and extensive experiments showcase its superior performance across a wide range of motion-relevant tasks.

\end{abstract}

% Note that keywords are not normally used for peerreview papers.
}

\maketitle

\IEEEdisplaynontitleabstractindextext
\IEEEpeerreviewmaketitle

\IEEEraisesectionheading{\section{Introduction}\label{sec:introduction}}
Human motion plays a pivotal role in diverse applications, including virtual reality, AR creation, and robotics \cite{xue2025human,liu2019ntu,zhu2023human,xu20213d}. Numerous studies focus on \textit{single-person motion comprehension}, \ie, deriving textural descriptions from 3D human motions, \cite{jiang2023motiongpt,wu2025mg,luo2024m} , and \textit{single-person motion generation}, \ie, generating 3D motions from diverse modalities including text \cite{zhang2024motiondiffuse,tevet2022human,guo2024momask}, music \cite{siyao2022bailando,tseng2023edge}, or poses \cite{zhang2024motiongpt}. 
Recent advances have expanded to more complex motion-pair tasks. These advanced tasks include \textit{interaction generation} (creating human-human interactions from text prompts) \cite{shafir2023human,tevet2022human,tanaka2023role}, 
\textit{motion reaction} (generating responsive motions to another agent's movements) \cite{chopin2023interaction,xu2024regennet,ghosh2024remos}, and \textit{motion editing} (refining existing motions based on modification instructions) \cite{fieraru2021aifit,goel2024iterative,athanasiou2024motionfix}.
Despite these advances, current methodologies treat these single- and paired-motion tasks as isolated problems. In practice, motion-pair tasks are intrinsically dependent on the underlying capabilities of single-motion comprehension and generation. This fragmentation underscores the critical need for a unified multimodal framework that can seamlessly integrate both single-motion and motion-pair tasks into a single, cohesive system.

Large language models (LLMs) have emerged as powerful general-purpose multitask systems, demonstrating impressive versatility across diverse domains \cite{touvron2023llama,touvron2023llama2,wu2023next}. Recent advances have successfully extended their application to human motion modeling, facilitating unified comprehension and generation of individual motion sequences. 
Notable frameworks such as MotionGPT \cite{jiang2023motiongpt} and $M^3$GPT \cite{luo2024m} employ a two-stage approach: initially,  a vector-quantized variational autoencoder (VQ-VAE) \cite{van2017neural}  converts continuous motion data into discrete token sequences. Subsequently,  a pretrained LLM \cite{raffel2020exploring}  learns to interpret this motion language, capturing its temporal dynamics and semantic correlations with textual descriptions.

\begin{figure}
\centering
\captionsetup{font={small}}
\includegraphics[width=0.9\linewidth]{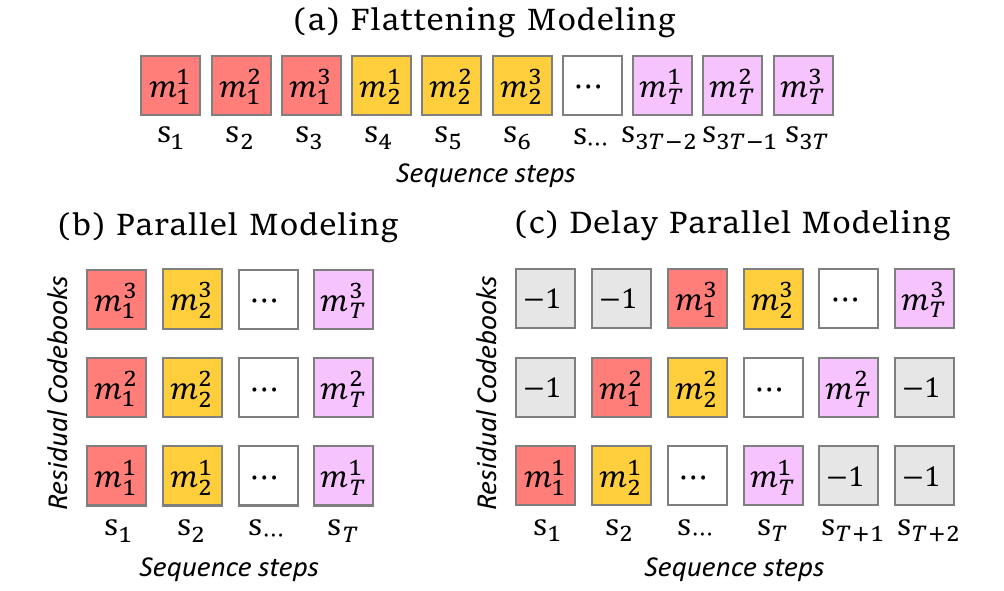}
\caption{Multi-stream motion token modeling strategies. We illustrate $3$ quantization levels as an example. Here, $m_t^l$ denotes the $t$-th token in the $l$-th motion stream. These motion tokens can be flattened or interleaved in various ways, resulting in a new sequence with either 1 or 4 parallel streams and steps $s_1,s_2,\dots,s_m$.  The special token $-1$ denotes empty positions within the pattern. }
\label{fig1}
\vspace{-10pt}
\end{figure}

Despite significant progress, current multitask approaches for human motion analysis exhibit three key limitations. \textbf{(i)} \textbf{Narrow Task Scope:} Existing studies focus primarily on modeling individual human motions (Tab. \ref{tabA1}), lacking unified frameworks capable of handling more complex tasks such as human-human interactions and motion editing. \textbf{(ii)} \textbf{Limited Representational Capacity}: Most approaches rely on basic vector quantization (VQ) \cite{raffel2020exploring}, employing a single, limited-size codebook. Consequently, diverse motion patterns must be encoded using a restricted set of predefined codes, leading to substantial information loss and insufficient representation of intricate human motion nuances. 
\textbf{(iii)} \textbf{Modality Interference}: Current approaches typically utilize a shared architecture to process both motion and language, resulting in representational conflicts between these inherently heterogeneous modalities. This leads to 
\textit{modality interference}: optimizing for one modality negatively impacts performance on the other, as their distinct statistical distributions and structural characteristics compete within the same parameter space. This phenomenon ultimately undermines the model's multitasking effectiveness.

To address these challenges, we introduce \textbf{MotionVerse}, a unified multimodal framework designed for the effective comprehension, generation, and editing of human motions. By leveraging the powerful generative capabilities of LLMs, MotionVerse seamlessly integrates multiple motion-related tasks (Tab. \ref{tabA1}) into a single cohesive system. Specifically, MotionVerse consists of two primary innovations.

\begin{figure*}
\centering
\captionsetup{font={small}}
\includegraphics[width=0.9\linewidth]{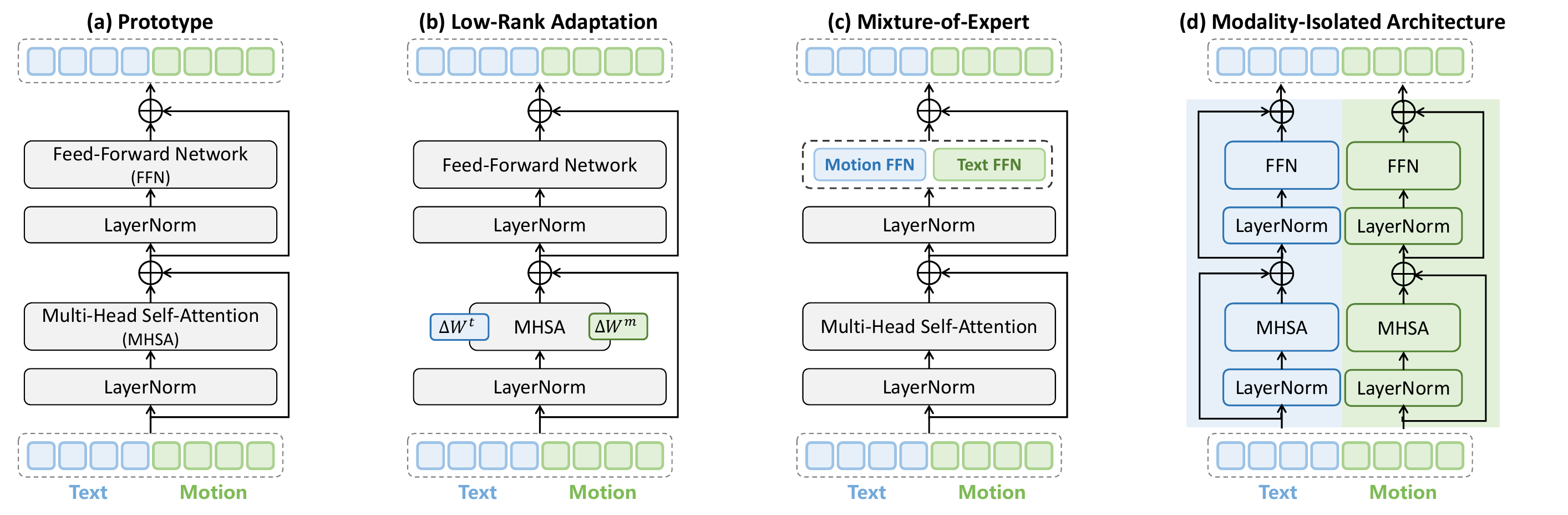}
\caption{Potential architecture variants of the Motion-Fusion module. (a) \textbf{Prototype}: Direct fine-tuning of a pretrained LLM. (b) \textbf{Low-Rank Adaptation} (LoRA): injecting low-rank matrices into multi-head self-attention layers. (c) \textbf{Mixture-of-Experts} (MoE): Duplicating each Feed-Forward Network (FFN) for the motion modality. (d) \textbf{Modality-Isolated Architecture} (MIS): Duplicating each transformer block for the motion modality.}
\label{fig2}
\vspace{-10pt}
\end{figure*}

First, we propose a unified modeling framework equipped with residual vector quantization (RVQ) \cite{lee2022autoregressive}. Specifically, (i) \textbf{Residual Vector Quantization for Motion Tokenization}: Different from vanilla vector quantization  \cite{jiang2023motiongpt,zhang2024motiongpt,luo2024m} that execute single-step quantization,  our framework employs RVQ to progressively encode motion. This is done by iteratively quantizing residuals, thereby systematically reducing errors. This process yields \textit{multi-stream motion tokens}, where the base stream captures fundamental motion characteristics, and subsequent streams encode residual errors sequentially. Such a hierarchical structure enables higher-quality motion comprehension and generation.  (ii) \textbf{Delay Pattern for Efficient Modeling}: A key challenge in RVQ is the simultaneous handling of multiple interdependent motion token streams by LLMs. A naive solution of unfolding all streams into a single extended sequence significantly increases computational demands (Fig. \ref{fig1} a). Alternatively, predicting multiple streams in parallel disregards the inherent residual dependencies (Fig. \ref{fig1} b). To mitigate this, we adopt a \textbf{delay pattern} strategy (Fig. \ref{fig1} c) inspired by \cite{copet2023simple}.
This strategy progressively delays the encoding of subsequent motion token streams. It effectively preserves inter-stream dependencies while maintaining a computational efficiency similar to single-stream modeling.

Second, we introduce a Motion-Fusion module with a dual-tower architecture that adapts a pretrained LLM to the motion modality. Motion-Fusion incorporates two modality-specific processing pathways within the LLM: a dedicated \textit{motion tower} and a \textit{text tower}, each maintaining independent parameters at every layer. This design allows the LLM to \textit{separately model} the distinct statistical properties and structural patterns of motion and text, thereby mitigating cross-modality interference. 
As shown in Fig. \ref{fig2}, we explore three implementations of the motion/text towers: (i) \textbf{Low-Rank Adaptation} (LoRA) \cite{hu2022lora}, which injects compact, low-rank matrices to enable efficient and lightweight adaptation; (ii) \textbf{Mixture-of-Experts} (MoE), where modality-specific expert modules are dynamically activated according to the input modality; and (iii) \textbf{Modality-Isolated Architecture} (MIS), which employs fully independent sub-models for motion and text, enforcing strict isolation to maximize specialization. 
Collectively, these strategies promote \textit{modality-specific specialization} within the LLM, significantly improving both the effectiveness and efficiency of multimodal and multitask learning.

To the best of our knowledge, MotionVerse is the first framework to unify $10$ core tasks covering motion comprehension, generation, reaction, and editing within a single cohesive architecture. Extensive experiments demonstrate that MotionVerse achieves competitive performance across various motion-related benchmarks, highlighting its broad effectiveness and versatility.

\section{Related Work}
\textbf{Generative Human Motion Modeling.} \
Human motion generation seeks to synthesize diverse and realistic motion sequences conditioned on multimodal inputs, such as text \cite{cen2024generating,guo2022generating,guo2024momask}, action \cite{petrovich2021action,guo2020action2motion}, incomplete motion sequences \cite{yuan2020dlow,zhang2021we,liu2022investigating}, or music \cite{siyao2022bailando,tseng2023edge}. Among these, text-to-motion generation has emerged as a prominent research direction, owing to the expressiveness and accessibility of natural language as a control signal. Synthesizing single-person motion has driven significant progress, enabled by the availability of large-scale motion capture datasets and advances in generative modeling techniques. 
Mainstream approaches can be broadly categorized into three families: diffusion-based methods \cite{zhang2024motiondiffuse,tevet2022human,raab2024monkey,kapon2024mas,raab2023single}, generative masked modeling \cite{pinyoanuntapong2024mmm,guo2024momask}, and autoregressive models \cite{zhang2023generating,lu2023humantomato}.

\begin{table*}[t]
\centering
\caption{Comparison of recent methods across various motion-relevant tasks. \textbf{M2T}: motion-to-text; \textbf{I-M2T}: interactive motion-to-text; \textbf{T2M}: text-to-motion; \textbf{I-T2M}: interactive text-to-motion; \textbf{M2M}: motion-to-motion that includes motion prediction and in-between; \textbf{I-M2M}: interactive motion-to-motion; \textbf{React}: motion reaction; \textbf{Edit}: motion editing. }
\begin{tabular}{l c c c c c c c c}
\toprule
\multirow{2}{*}{Methods} & \multicolumn{2}{c}{Comprehension} & \multicolumn{4}{c}{Generation} &Reaction & Editing \\
\cmidrule(lr){2-3}  \cmidrule(lr){4-7} \cmidrule(lr){8-8} \cmidrule(lr){9-9} 
& M2T & I-M2T & T2M & I-T2M & M2M & I-M2M &React & Edit \\
\midrule
MotionGPT  \cite{jiang2023motiongpt}  & $\checkmark$ & $\times$  &$\checkmark$ &$\times$ & $\checkmark$ &$\times$& $\times$& $\times$ \\
MotionGPT-13B \cite{zhang2024motiongpt} & $\times$ & $\times$  &$\checkmark$ &$\times$ & $\checkmark$ &$\times$& $\times$& $\times$ \\
M$^3$GPT \cite{luo2024m}   & $\checkmark$ & $\times$  &$\checkmark$ &$\times$ & $\checkmark$ &$\times$& $\times$& $\times$ \\
MG-MotionLLM \cite{wu2025mg} & $\checkmark$ & $\times$  &$\checkmark$ &$\times$ & $\checkmark$ &$\times$& $\times$& $\times$ \\
AvatarGPT \cite{zhou2024avatargpt} & $\checkmark$ & $\times$  &$\checkmark$ &$\times$ & $\checkmark$ &$\times$& $\times$& $\times$ \\
MotionLLM \cite{wu2024motionllm} &  $\checkmark$ & $\times$  &$\checkmark$ &$\checkmark$ & $\checkmark$ &$\times$& $\times$& $\times$ \\
\midrule
MotionVerse (Ours) &$\checkmark$ & $\checkmark$  &$\checkmark$ &$\checkmark$ & $\checkmark$ &$\checkmark$& $\checkmark$& $\checkmark$ \\
\bottomrule
\end{tabular}
\label{tabA1}
\end{table*}

\textbf{Diffusion-based methods} generate motion sequences by progressively denoising an initial noise signal through a scheduled diffusion process. For example, MDM \cite{tevet2022human} employs a transformer-based diffusion model conditioned on  CLIP-extracted text features. MotionDiffuse \cite{zhang2024motiondiffuse} further achieves part-independent control using fine-grained textual descriptions.  Recent works such as MoMo \cite{raab2024monkey}, MAS \cite{kapon2024mas}, and SinMDM \cite{raab2023single} demonstrate strong generalization to out-of-domain motion data.  To enhance controllability, A-MDM \cite{shi2024interactive} and CAMDM \cite{chen2024taming}  introduce autoregressive diffusion frameworks that support interactive control mechanisms.
\textbf{Generative masked modeling} reconstructs complete motion sequences from partially masked inputs. For instance,  MMM \cite{pinyoanuntapong2024mmm} employs a conditional masked motion transformer guided by textual descriptions.  
MoMask \cite{guo2024momask} enhances generation quality by integrating residual vector quantization \cite{lee2022autoregressive} with masked modeling.
\textbf{Autoregressive approaches} discretize continuous motion into tokenized sequences and model them in an autoregressive manner. T2M-GPT \cite{zhang2023generating} combines a VQ-VAE  \cite{van2017neural} with a GPT-based transformer \cite{vaswani2017attention} for conditional motion generation. Building on this foundation, HumanTOMATO \cite{lu2023humantomato} incorporates a hierarchical VQ-VAE and GPT modules to capture fine-grained body and hand motion dynamics.  

\vspace{0.5\baselineskip}

\noindent
\textbf{Generative Human-Human Interaction Modeling.} \
Most research on human interaction modeling has focused on two-person interactions, with efforts concentrated on two tasks: interaction generation and reaction generation. 

In \textbf{Interaction Generation}, the focus is on synthesizing motion trajectories for two individuals at the same time. For example,  ComMDM \cite{shafir2023human} connects two  single-person motion diffusion models \cite{tevet2022human} with a lightweight coordination module to enable joint motion synthesis, even with limited interaction data.
Other approaches, like RIG \cite{tanaka2023role} and InterGen \cite{liang2024intergen}, use interaction-aware diffusion architectures that simultaneously denoise the motion sequences of both participants, conditioned on each other's latent representations. InterMask \cite{javed2024intermask} employs a generative masked modeling framework to collaboratively model the token sequences of both interacting individuals.   To improve interaction fidelity, several approaches \cite{cai2024digital,ruiz2024in2in} have incorporated auxiliary single-person motion annotations and leveraged external motion datasets. More recent research \cite{shan2024towards,fan2024freemotion} has also begun to move beyond two-person interactions to address the complexities of multi-party interactions, with models designed to handle three or more individuals.

In \textbf{Reaction Generation}, the motion of one individual is generated in response to the actions of another. 
For example,  InterFormer \cite{chopin2023interaction} uses a transformer-based architecture with spatiotemporal attention to model responsive human behaviors. Alternatively, ReGenNet \cite{xu2024regennet} employs a diffusion-based framework with a transformer decoder to synthesize these reactive motions.  Extending these capabilities, ReMoS \cite{ghosh2024remos} focuses on generating realistic hand motions for the reactor, facilitating realistic hand-centric interactions. To ensure physical plausibility, \cite{liu2024physreaction} incorporates a learned dynamics module within an imitation learning framework.  Despite the effectiveness of these approaches, most of them treat human-human interactions in isolation. A key challenge remains:  developing a unified framework that can model both individual motion generation and interactive behaviors seamlessly. 

\vspace{0.5\baselineskip}

\noindent
\textbf{Human Motion Editing.} \
Existing motion editing methods can be broadly categorized into three paradigms. (1) Style-based editing, which transfers stylistic attributes between motions while preserving their underlying structural dynamics \cite{aberman2020unpaired, mason2022real}; (2) Part-based editing, which enables localized manipulation of specific body regions using techniques such as text-guided diffusion models \cite{tevet2022human}, manual region selection \cite{zhang2024motiondiffuse, kim2023flame}, or LLM-generated editing instructions \cite{huang2024controllable}; (3)  Heuristic-based editing, which relies on rule-based strategies, including motion grammar for exercise routines \cite{fieraru2021aifit}, and LLM-assisted joint refinement through diffusion-based inpainting \cite{goel2024iterative}.  Despite these advances, most existing approaches remain tailored to narrow editing scenarios or rely heavily on handcrafted heuristics.  To address these limitations,  MotionFix \cite{athanasiou2024motionfix} proposes a more flexible framework that leverages a conditional diffusion model to directly generate edited motions, conditioned jointly on the source motion and textural editing instructions. 

\vspace{0.5\baselineskip}

\noindent
\textbf{Language Models and Multimodal.} \ 
Powered by large-scale datasets and model scaling, large language models such as T5 \cite{raffel2020exploring}, LLaMA \cite{touvron2023llama}, and Vicuna \cite{chiang2023vicuna}  have exhibited remarkable capabilities in language understanding and generation. Building on this foundation, recent research has extended LLMs to support multimodal inputs,  giving rise to multimodal large language models (MLLMs). For instance,  AnyGPT \cite{zhan2024anygpt} integrates LLaMA-2 \cite{touvron2023llama2} into an any-to-any multimodal framework, while NExT-GPT \cite{wu2023next} augments Vicuna   \cite{chiang2023vicuna} with modality-specific adapters and diffusion-based decoders, thereby enabling flexible interactions across text, images, video, and audio.

More recently, the application of LLMs has expanded into the domain of human motion. 
For example, \cite{zhang2024motiongpt} leverages LLaMA \cite{touvron2023llama} to develop a versatile motion generation framework, while \cite{jiang2023motiongpt} employs T5 \cite{raffel2020exploring} to construct a unified motion-language model capable of both understanding and generating motions. Building on this foundation, 
$M^3$GPT \cite{luo2024m} enables bidirectional translation between music and dance choreography, and  \cite{chen2024language} integrates a body-part VQ-VAE with T5 to support multimodal conditioning on both text and speech. Furthermore, MG-MotionLLM \cite{wu2025mg} introduces multi-granularity modeling for motion understanding and generation, while AvatarGPT \cite{zhou2024avatargpt} proposes an all-in-one framework that unifies motion understanding, planning, and generation within a single model. 
MotionLLM \cite{wu2024motionllm} further
demonstrates the generalization and flexibility of LLMs by extending single-person motion generation to more complex multi-person motion scenarios.

Despite these advancements, most approaches remain focused on modeling individual motions, with limited capacity to capture complex human-human interactions or support motion editing. In addition, many rely on basic vector quantization techniques and modality-agnostic architectures, which overlook the distinct structural and semantic properties of different input modalities (\eg, motion vs. text). Consequently, their representational power and ability to generalize across diverse motion-related tasks remain significantly constrained.

\begin{figure*}
\centering
\captionsetup{font={small}}
\includegraphics[width=0.9\linewidth]{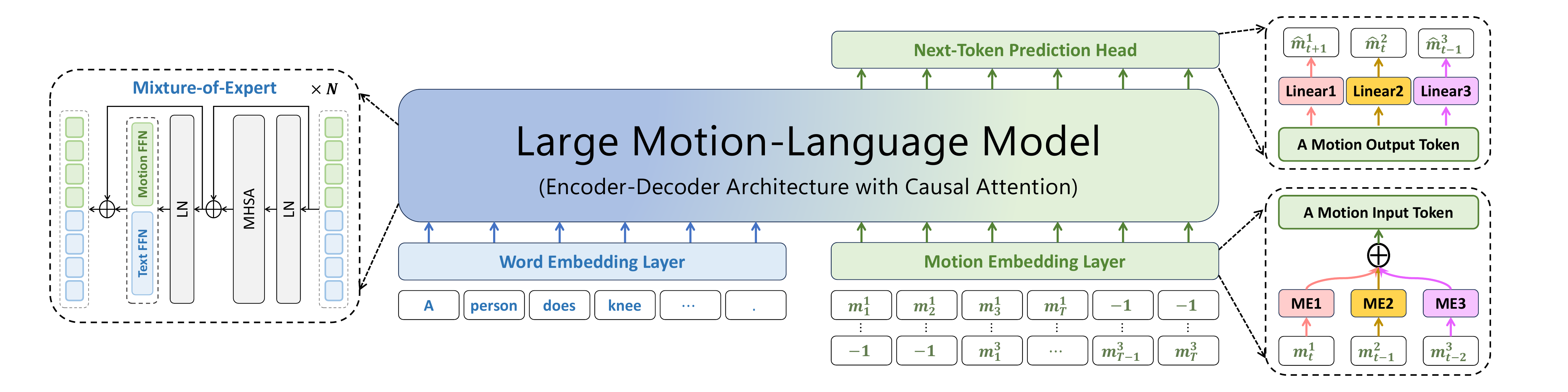}
\caption{Architecture of the proposed MotionVerse framework: Integrating Textural and Multi-stream Motion Tokens via Modality-Aware Autoregressive Transformers (Three-Stream Example).} %ME denotes a Motion Embedding layer. }
\label{fig3}
\vspace{-10pt}
\end{figure*}

\section{Method}
To endow large language models (LLMs) with the capability to comprehend, generate, react to, and edit human motion, we propose a unified framework named \textbf{MotionVerse}. The overall architecture is illustrated in Fig. \ref{fig3}.
MotionVerse comprises three main components: (1) a residual quantizer that discretizes motion sequences into multi-stream tokens while faithfully preserving fine-grained motion dynamics; (2) a parallel, stream-aware LLM backbone that jointly models multiple streams of motion tokens through a \textit{temporal delay} mechanism,  enabling high-quality motion generation with negligible computational overhead; and (3) a Motion-Fusion module that further strengthens the LLM's capacity for motion understanding and generation. To support versatile motion-centric tasks, MotionVerse is optimized via a three-stage training pipeline that includes motion tokenizer training, motion-text alignment pretraining, and multitask instruction tuning.

\subsection{Motion Residual VQ-VAE}
\textbf{Residual Vector Quantization (RVQ).} \ 
The RVQ process employs a set of $L$ codebooks, denoted as $\left\{\mathcal{B}^l\right\}_{l=1}^L$. Each codebook $\mathcal{B}^l$ can be formulated as $\left\{k, \mathbf{e}^l\left(k\right)\right\}_{k\in \left[K\right]}$, which consists of the pairs of a code $k$ and its code embedding $\mathbf{e}^l\left(k\right) \in \mathcal{R}^d$, where $K$ is the codebook size and $d$ is the dimensionality of code embeddings. Given a vector $\boldsymbol{z}\in\mathcal{R}^{d}$, RVQ represents $\boldsymbol{z}$ as an \textit{ordered} $L$ codes:
\begin{equation}
\mathcal{RQ}\left(\boldsymbol{z}; \mathcal{B}^1, \dots, \mathcal{B}^L\right)=\left(k^1,\dots, k^L\right),
\end{equation}
where $k^l$ is the discrete code of $\boldsymbol{z}$ at depth $l$. Specifically, starting with %$0$-th residual 
$\boldsymbol{r}^1=\boldsymbol{z}$, RVQ recursively computes $k^l$ (code of the residual $\boldsymbol{r}^{l}$), and the next residual $\boldsymbol{r}^{l+1}$ as:
\begin{equation}
\begin{split}
k^l &= \mathcal{Q}\left(\boldsymbol{r}^{l}; \mathcal{B}^l\right),\\
\boldsymbol{r}^{l+1} &= \boldsymbol{r}^{l} - \mathbf{e}^l(k^l),  \enspace \text{for} \ l=1,\dots,L.
\label{eq2}
\end{split}
\end{equation}
Here, $\mathcal{Q}\left(\boldsymbol{r}^{l}; \mathcal{B}^l\right)=\mathop{\arg\min}\limits_{k\in[K]}\left\|\boldsymbol{r}^{l} - \mathbf{e}^l\left(k\right)\right\|_2^2$ denotes the vector quantization of $\boldsymbol{r}^{l}$, which selects the code within codebook $\mathcal{B}^l$ whose embedding is nearest to  $\boldsymbol{r}^{l}$. 
We define the approximation of the residual $\boldsymbol{r}^{l}$ as 
$\boldsymbol{\hat{r}}^l=\mathbf{e}^l(k^l)$. The final quantized representation of original vector $\boldsymbol{z}$ is then given by  $\boldsymbol{\hat{z}}=\sum_{l=1}^L\boldsymbol{\hat{r}}^{l}$.

\vspace{0.5\baselineskip}

\begin{figure}
\centering
\captionsetup{font={small}}
\includegraphics[width=0.8\linewidth]{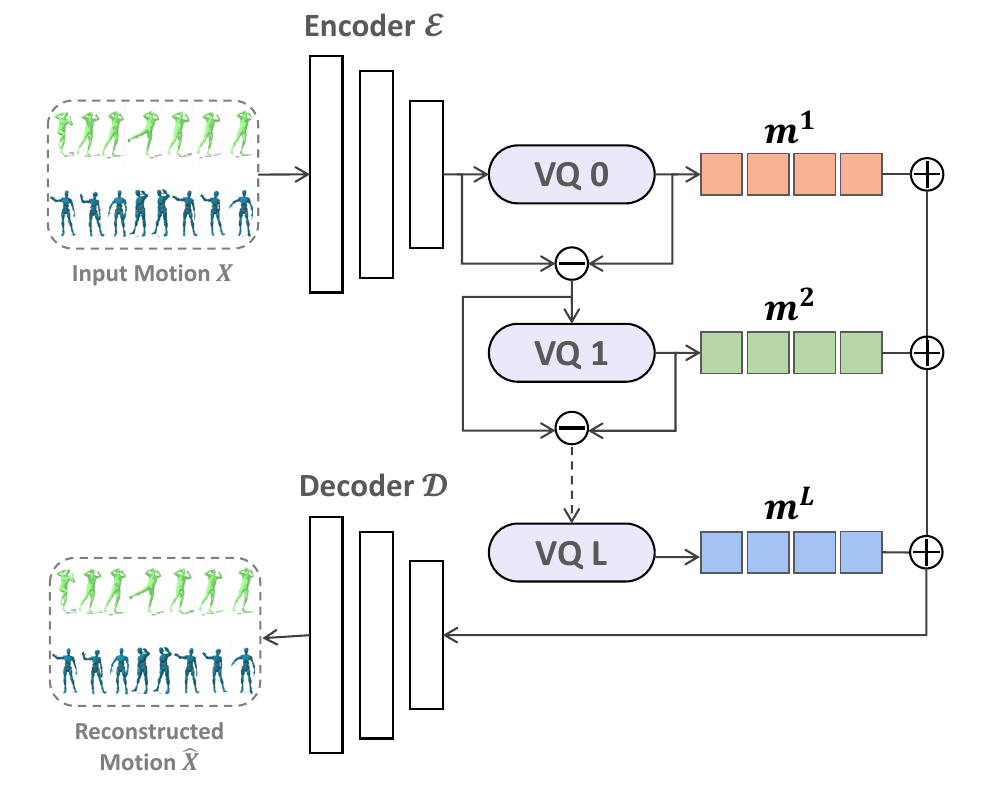}
\caption{Architecture of Residual Vector Quantization Variation Autoencoder (RVQ-VAE), which discretizes continuous motions into $L$ token sequences.}
\label{fig4}
\vspace{-10pt}
\end{figure}
\noindent
\textbf{Motion RVQ-VAE.} \ 
To represent motion as discrete semantic tokens, we build a 3D motion tokenizer based on RVQ following \cite{guo2024momask}. 
As shown in Fig. \ref{fig4}, the motion tokenizer consists of a motion encoder $\mathcal{E}$ and a motion decoder $\mathcal{D}$, along with $L$ codebooks $\left\{\mathcal{B}^l\right\}_{l=1}^L$. 
Formally, given a 3D motion sequence $\mathbf{X}\in\mathcal{R}^{T_o\times d_o}$, where $T_o$ is the sequence length and $d_o$ is the dimensionality of each pose frame, the motion encoder $\mathcal{E}$, composed of several 1D convolutional layers, projects $\mathbf{X}$ to a latent feature map $\mathbf{Z} \in\mathcal{R}^{T\times d}$ with downsampling ratio of $T_o/T$ and latent dimension $d$. By applying RVQ to each feature vector in $\mathbf{Z}$,  the motion tokenizer quantizes it into a stacked map of discrete codes, producing a code map $\mathbf{M} \in \left[K\right]^{T \times L}$ and extracting $\mathbf{\widehat{R}}^l \in \mathcal{R}^{T \times d}$, which is the quantized residual feature map at level $l$ such that:
\begin{equation}
\begin{split}
\mathbf{M}\left[t\right] = \mathcal{RQ}\left(\mathbf{Z}\left[t\right]; \mathcal{B}^1, \dots, \mathcal{B}^L\right) , \ \mathbf{\widehat{R}}^l\left[t\right]  = \mathbf{e}^l\left(\mathbf{M}\left[t,l\right]\right),
\end{split}
\end{equation}
where $\mathbf{M}\left[t\right]$ denotes the $t$-th row of $\mathbf{M}$, analogously for $\mathbf{Z}\left[t\right]$ and  $\mathbf{\widehat{R}}^l\left[t\right]$. 
Then the quantized feature map is obtained as $\mathbf{\widehat{Z}}=\sum_{l=1}^L\mathbf{\hat{R}}^l$. Finally, the motion decoder $\mathcal{D}$, which consists of several 1-D deconvolutional layers, reconstructs the input motion from $\mathbf{\hat{Z}}$ as $\mathbf{\widehat{X}}=\mathcal{D}\left(\mathbf{\widehat{Z}}\right)$.

 Following \cite{lee2022autoregressive,guo2024momask}, the residual VQ-VAE is trained by a motion reconstruction loss combined with a commitment loss at each quantization layer: 
\begin{equation}
\mathcal{L}_{\mathrm{rvq}}=\left\|\mathbf{X}-\mathbf{\widehat{X}}\right\|_1 + \beta \sum_{l=1}^L\left\|\mathbf{R}^l - \mathrm{sg}\left[\mathbf{\widehat{R}}^l\right]\right\|_2^2,
\label{eq4}
\end{equation}
where $\mathrm{sg}\left[\cdot\right]$ denotes the stop-gradient operation, and $\beta$ is a hyperparameter controlling the commitment loss. 

After training the motion tokenizer, a motion sequence $\mathbf{X}$ can be represented as $L$ discrete motion token sequences, denoted as  $\mathbf{M}=\left[\boldsymbol{m}^1, \dots, \boldsymbol{m}^L\right]$, where  $\boldsymbol{m}^l=\mathbf{M}\left[:,l-1\right]$ corresponds to the token sequence at the $l$-th stream. Among these $L$ sequences, the first (\ie, base) stream $\boldsymbol{m}^1$ captures the most prominent information, while the impact of subsequent residual streams $\left(\boldsymbol{m}^2, \dots, \boldsymbol{m}^L\right)$ gradually diminishes with increasing $l$.

\subsection{Parallel-Stream Aware LLM Backbone}
\label{sec:3.2}
% \textbf{Unified Multimodal Language Model.} \ 
\textbf{Expanding Vocabulary.} \ 
To incorporate discrete motion tokens of multiple streams into a pre-trained LLM, we expand the original text vocabulary $\mathcal{V}_t$ with a hierarchical motion vocabulary $\left\{\mathcal{V}^1_m, \dots, \mathcal{V}^L_m\right\}$, where each sub-vocabulary  $\mathcal{V}^l_m$ preserves the ordering of its corresponding motion codebook $\mathcal{B}^l$. This yields a unified text-motion vocabulary $\mathcal{V}=\left\{\mathcal{V}_t, \mathcal{V}^1_m, \dots, \mathcal{V}^L_m\right\}$, enabling the formulation of  diverse motion-related tasks under a general sequence-to-sequence framework. 
To accommodate this enlarged vocabulary, we extend both the embedding layer and the output prediction layer of LLM, with the newly added parameters initialized randomly.
Within this framework, input and output sequences can seamlessly consist of natural language tokens, motion tokens, or a combination of both, depending on the task at hand.
This design naturally enables the autoregressive LLM to unify motion comprehension, generation, and editing within a single model. 

\vspace{0.5\baselineskip}

\begin{table}[t]
\centering
\caption{Comparison of Multi-stream Motion Tokens Modeling Strategies. \textbf{Intra-stream}: Capability to model temporal dependencies within each stream. \textbf{Inter-stream}: Capability to model hierarchical residual dependencies across parallel streams. \textbf{Complexity}: Attention complexity within the LLM. Here, $T$ denotes the motion sequence length, and $L$ is the number of quantization layers. }
\begin{tabular}{l c c c c c c c c}
\toprule
Strategy & Intra-stream & Inter-stream & Complexity\\
\midrule
Flattening & $\checkmark$ & $\checkmark$ & $O\left(T^2L^2\right)$\\
Parallel & $\checkmark$ & $\times$ & $O\left(T^2\right)$\\
Delay Parallel & $\checkmark$ & $\checkmark$ & $O\left(\left(T+L-1\right)^2\right)$ \\
\bottomrule
\end{tabular}
\label{tabA2}
\end{table}
      
\noindent
\textbf{Multi-Stream Motion Tokens Modeling.} \  
Unlike prior motion-aware LLM approaches that encode motion as a single token stream \cite{jiang2023motiongpt,zhang2024motiongpt,luo2024m}, we employ a Residual Vector Quantization VAE (RVQ-VAE) to discretize motion into multiple parallel token streams $\left\{\boldsymbol{m}^1, \dots, \boldsymbol{m}^L\right\}$. For a specific motion token $m^l_t$, the LLM is required to model two distinct types of dependencies: (1) \textbf{intra-stream dependencies}, capturing the temporal dynamics within each stream by attending to the preceding tokens, $\boldsymbol{m}^l_{<t}$; and (2) \textbf{inter-stream dependencies}, capturing hierarchical relationships across streams, as each stream encodes residual information conditioned on its predecessors, modeling relationships with $\boldsymbol{m}^{<l}_{t}$.

To address this modeling challenge, we systematically explore three distinct strategies: Flattening, Parallel, and Delay Parallel. A comparison of these strategies is presented in Tab.~\ref{tabA2}. 

%\vspace{0.5\baselineskip} 

\textbf{{Flattening Modeling.}} \ As shown in Fig. \ref{fig1} (a), a simple solution is to flatten the $L$ parallel token sequences into a single token stream $U$ of length $T\times L$. 
\begin{equation}
U = \left[m^1_1, m^2_1, \dots, m^L_1,  \dots, m^1_T, m^2_T, \dots, m^L_T \right], 
\end{equation}
where $m^l_t$ denotes the $t$-th token in the $l$-th motion stream $\boldsymbol{m}^l$. At each position in the flattened sequence, LLM applies standard causal self-attention to model the conditional probability: 
\begin{equation}
p\left(m^l_t|\boldsymbol{m}^{<l}_{t}, \boldsymbol{m}^{1}_{<t}, \dots, \boldsymbol{m}^{L}_{<t} \right). 
\label{eq6}
\end{equation}
The formulation allows the model to capture: (1) 
\textit{Intra-stream temporal dependencies} via $\boldsymbol{m}^l_{<t}$, \ie, earlier-timestep tokens within the same stream; (2) 
\textit{Inter-stream dependencies} via  $\boldsymbol{m}^{<l}_t$, \ie, tokens from preceding streams at the same timestep. 
However, this flattening strategy incurs substantial computational overhead. The sequence length increases from $T$ to $TL$, escalating the attention complexity from $O\left(T^2\right)$ to $O\left(T^2L^2\right)$. 
This computational growth limits the scalability of the flattening strategy for long motion sequences.

\vspace{0.5\baselineskip} 

\textbf{{Parallel Modeling.}} \
As shown in Fig. \ref{fig1} (b), an alternative strategy is to encode and predict multiple motion streams in parallel. However, this design poses a fundamental challenge: LLMs are inherently designed for \textit{sequential} token processing, which leads to a paradigm mismatch when modeling parallel streams. 
To address this issue, we introduce two key architectural innovations that enable the LLM to effectively perceive and generate motion tokens across parallel streams:

(1) \textbf{Motion Perception}:  To enable parallel encoding of multi-stream motion tokens, we introduce a set of $L$  motion embedding layers, denoted as $\left\{f^l\left(\cdot\right)\right\}_{l=1}^L$, where each embedding layer $f^l$ 
independently processes the corresponding motion token stream $\boldsymbol{m}^l$. The final motion representation is obtained by aggregating the outputs of all embedding layers:
\begin{equation}
\mathrm{Z}_{\mathrm{enc}} = \sum_{l=1}^L f^l\left(\boldsymbol{m}^l\right).
\label{eq7}
\end{equation}
This aggregated motion representation $\mathrm{Z}_{\mathrm{enc}}$ is then fed into the LLM backbone for subsequent processing.

(2) \textbf{Motion Generation}: To enable parallel generation of multi-stream motion tokens, we introduce a set of $L$   motion prediction layers $\left\{g^l\left(\cdot\right)\right\}_{l=1}^L$, where each layer $g^l$ is responsible for generating motion tokens at the corresponding level $l$. These prediction layers operate in parallel, each taking the shared latent representation from the LLM backbone as input, and decoding it into one of the $L$ distinct motion streams:
\begin{equation}
\boldsymbol{\widehat{m}}^l = g^l\left(\mathrm{Z}_{\mathrm{dec}}\right), \enspace \text{for} \ l=1,\dots,L.
\label{eq8}
\end{equation}
where $\mathrm{Z}_{\mathrm{dec}}$ denotes the shared latent representation produced by the LLM. 

According to Eq. \ref{eq7} and \ref{eq8}, at each timestep, the LLM employs causal self-attention to model the conditional probability:
\begin{equation}
p\left(m^1_t, m^2_t\dots, m^{L}_t|\boldsymbol{m}^{1}_{<t}, \boldsymbol{m}^{2}_{<t}, \dots, \boldsymbol{m}^{L}_{<t}\right).
\label{eq9}
\end{equation}
This mechanism preserves the original motion sequence length $T$ during modeling, leading to a computationally efficient attention complexity of $O\left(T^2\right)$. However, this efficiency comes at the cost of reduced modeling accuracy, as it neglects inter-stream dependencies. Specifically, as shown in Eq. \ref{eq9}, while parallel modeling effectively captures \textit{intra-stream temporal dependencies} by conditioning on past tokens within each stream, it inherently fails to account for \textit{inter-stream dependencies}, \ie, the correlations between the current token $m^l_t$ and tokens from preceding streams at the same timestep ($\boldsymbol{m}^{<l}_t$). 
This limitation is particularly critical in the context of residual vector quantization, where each subsequent stream encodes residual information relative to its preceding streams (Eq. \ref{eq2}), thereby introducing a hierarchical dependency across streams. Consequently,  parallel modeling is fundamentally suboptimal for capturing such hierarchical inter-stream dependencies.

\vspace{0.5\baselineskip} 

\textbf{{Delay Parallel Modeling.}}
We further explore a delay parallel modeling approach. As shown in Fig. \ref{fig1} (c), we introduce a \textit{temporal delay} to the motion tokens of each residual stream $\boldsymbol{m}^l$ \cite{copet2023simple,kharitonov2021text}. This delay creates a new sequence$ \boldsymbol{\widetilde{m}}^l$ of length $T+L-1$:
\begin{equation}
\boldsymbol{\widetilde{m}}^l = \left[\underbrace{-1, \dots, -1}_{l-1}, \boldsymbol{m}^l, \underbrace{-1, \dots, -1}_{L-l}\right], \  l=1,\dots,L.
\label{eq10}
\end{equation}
Here, $-1$ denotes a special padding token.
Given the delayed multi-stream  sequences $\left\{\boldsymbol{\widetilde{m}}^1, \dots, \boldsymbol{\widetilde{m}}^L\right\}$, we perform motion perception and generation effectively by applying Eq.~\ref{eq7} (substituting $\boldsymbol{m}^l \rightarrow \boldsymbol{\widetilde{m}}^l$) and Eq.~\ref{eq8} (substituting $\boldsymbol{\widehat{m}}^l \rightarrow \boldsymbol{\widehat{\widetilde{m}}}^l$).
The delay operation enables the LLM to leverage standard  causal self-attention to model the conditional probability at each timestep:
\begin{equation}
p\left(m^1_t, m^2_{t-1},\dots,m^L_{t-L+1}|\boldsymbol{m}^1_{<t}, \boldsymbol{m}^2_{<t-1}, \dots, \boldsymbol{m}^L_{<t-L+1}\right).
\label{eq11}
\end{equation}
As shown in Eq. \ref{eq11}, for a specific motion token $m^l_{t}$, the delayed parallel mechanism effectively captures both \textit{Intra-stream temporal dependencies} via $\boldsymbol{m}^l_{<t}$, and \textit{Inter-stream residual dependencies} via $\boldsymbol{m}^{<l}_{t}$.

As shown in Eq. \ref{eq10}, the delay operation increases the length of the input motion sequence from $T$ to $T+L-1$. This leads to an attention complexity of $O\left(\left(T+L-1\right)^2\right)$. However, since the motion sequence length $T$ is typically much larger than the number of residual quantization levels $L$, the additional computational overhead introduced by the delay operation is negligible, especially for long motion sequences. Notably, this complexity remains substantially lower than the $O\left(T^2L^2\right)$ incurred by the Flattening strategy, making Delay Parallel modeling a more scalable and efficient alternative.

Based on the above analysis, the Delayed Parallel modeling effectively captures both intra-stream and inter-stream dependencies while incurring only negligible additional computational overhead. Consequently, our MotionVerse framework adopts the Delayed Parallel strategy within the LLM backbone.

\subsection{Motion-Fusion: Empowering LLMs with Motion Modeling Capabilities}
In this section, we introduce Motion-Fusion, a module designed to adapt pretrained LLMs for motion-related tasks. To mitigate interference between heterogeneous modalities,  namely motion and text, Motion-Fusion incorporates two distinct processing pathways within the LLM: a \textbf{motion tower} and a \textbf{text tower}, each specialized for its respective modality. 
We systematically explore three architectural variants for implementing these modality-specific towers: \textbf{Low-Rank Adaptation (LoRA)}, \textbf{Mixture-of-Expert (MoE)}, and \textbf{Modality-Isolated Architecture (MIS)}. 

\vspace{0.5\baselineskip} 

\noindent
\textbf{Prototype.} \ 
The Transformer block serves as the fundamental computational unit in LLMs, typically consisting of three key subcomponents: Layer Normalization (LN), Multi-Head Self-Attention (MHSA), and Feed-Forward network (FFN). We first examine a prototype baseline model \cite{jiang2023motiongpt,zhang2024motiongpt,luo2024m}, which applies identical Transformer blocks to both motion and text modalities within the LLM architecture, as illustrated in  Fig. \ref{fig2} (a).

Formally, let the tokenized input embeddings be denoted as $\boldsymbol{x} = \left(\boldsymbol{x}_1, \dots, \boldsymbol{x}_n\right)$, where each $\boldsymbol{x}_i \in \mathbb{R}^{d_k}$ corresponds to the embedding of either a motion or text token. The operations within a standard Transformer block can be formally expressed as follows: 
\begin{equation}
\begin{split}
\boldsymbol{h} &= \boldsymbol{x} + \mathrm{MHSA}\left(\mathrm{LN}\left(\boldsymbol{x}, \theta_{\mathrm{ln1}}\right); \theta_{\mathrm{attn}} \right), \\
\boldsymbol{x'} &= \boldsymbol{h} + \mathrm{FFN}\left(\mathrm{LN}\left(\boldsymbol{x}, \theta_{\mathrm{ln2}}\right); \theta_{\mathrm{ffn}}\right). 
\label{eq12}
\end{split}
\end{equation}
Here, $\mathrm{LN}\left(\cdot\right)$ denotes the layer normalization \cite{zhang2019root}, and $\mathrm{MHSA}\left(\cdot\right)$ denotes the multi-head self-attention mechanism \cite{vaswani2017attention}. MHSA with $h$ heads is computed as: 
\begin{equation}
\begin{split}
&\mathrm{MHSA}\left(\boldsymbol{x}; \theta_{\mathrm{attn}} \right)=\mathrm{Concat}\left(\mathrm{head_1},\dots,\mathrm{head_h}\right)W^O, \\
&\mathrm{head_i} = \mathrm{softmax}\left(\boldsymbol{Q}_i\boldsymbol{K}_i^T/\sqrt{d_k}\right)\boldsymbol{V}_i, \\
& \boldsymbol{Q} = \boldsymbol{x}W_{Q}, \quad \boldsymbol{K} = \boldsymbol{x}W_{K}, \quad \boldsymbol{V} = \boldsymbol{x}W_{V}.
\label{eq13}
\end{split}
\end{equation}
Here, %$\theta_{\mathrm{attn}}=\left\{W_Q,W_K,W_V,W_O\right\}$, 
$W_{Q}$, $W_{K}$, $W_{V}$, $W_{O} \in \mathbb{R}^{d_k \times d_k}$ are learned projection matrices for the query, key, value, and output, respectively. The $i$-th head's submatrix, for example $Q_i$ is extracted as $Q_i=Q\left[:, i\frac{d_k}{h}:\left(i+1\right)\frac{d_k}{h}\right]$, with analogous definitions for $K_i$ and $V_i$.

In Eq. \ref{eq12}, $\mathrm{FFN}\left(\cdot\right)$ is a two-layer feed-forward network with a non-linear activation function:
\begin{equation}
\mathrm{FFN}\left(\boldsymbol{x}; \theta_{\mathrm{ffn}}\right) = W_2 \cdot \sigma\left(W_1 \boldsymbol{x} + b_1\right) + b_2,
\label{eq14}
\end{equation}
where $W_1\in\mathbb{R}^{d_{f}\times d_k}$, $W_2\in\mathbb{R}^{d_{k}\times d_f}$ are learnable projection matrices, $b_1$, $b_2$ are bias terms, and $\sigma$ is a non-linear activation function. 

\vspace{0.5\baselineskip} 

\noindent
\textbf{Low-RanK Adaptation (LoRA).} \ 
LoRA is a parameter-efficient fine-tuning method that injects trainable low-rank  matrices into pre-trained weights to enable efficient adaptation. Inspired by this, we introduce \textbf{Motion-LoRA} and \textbf{Text-LoRA}, two modality-specific adaptation mechanisms designed to independently model motion and text tokens, as illustrated in  Fig. \ref{fig2} (b).
Formally, given a pre-trained weight matrix $W\in\mathbb{R}^{d_k\times d_k}$, we augment it with two distinct low-rank adaptations, yielding a motion-specific weight $W^{\mathrm{lora}\text{-}\mathrm{m}}$ and a text-specific weight $W^{\mathrm{lora}\text{-}\mathrm{t}}$. This process can be formulated as: 
\begin{equation}
\begin{split}
&W^{\mathrm{lora}\text{-}\mathrm{m}},W^{\mathrm{lora}\text{-}\mathrm{t}} = \mathrm{LoRA}_{\mathrm{MT}}\left(W\right), \quad \text{where}, \\
&W^{\mathrm{lora}\text{-}\mathrm{m}} = W + \alpha \Delta W^m, \quad \Delta W^m = B^mA^m, \\
&W^{\mathrm{lora}\text{-}\mathrm{t}} = W + \alpha \Delta W^t, \quad \Delta W^t = B^tA^t.
\label{eq15}
\end{split}
\end{equation}
Here, $B^m,B^t\in\mathbb{R}^{d_k\times r}$ and $A^m, A^t \in \mathbb{R}^{r\times d_k}$ are trainable low-rank matrices for motion and text, respectively.  The rank $r \ll d_k$ controls the adaptation capacity, and $\alpha$ is a scaling factor. 
Following common practice, we integrate Motion-LoRA and Text-LoRA  into the projection matrices of the self-attention layers (Eq. \ref{eq13}). Specifically, the standard projection matrices $\left\{W_Q,W_K,W_V,W_O\right\}$ are each augmented with Motion-LoRA and Text-LoRA modules, enabling the model to adaptively process motion and text tokens through separate low-rank adaptations: 
\begin{equation}
\begin{split}
&\theta^{\mathrm{lora}\text{-}\mathrm{m}}_{\mathrm{attn}}=\left\{W^{\mathrm{lora}\text{-}\mathrm{m}}_{Q},W^{\mathrm{lora}\text{-}\mathrm{m}}_{K},W^{\mathrm{lora}\text{-}\mathrm{m}}_{V},W^{\mathrm{lora}\text{-}\mathrm{m}}_O\right\},\\
&\theta^{\mathrm{lora}\text{-}\mathrm{t}}_{\mathrm{attn}}=\left\{W^{\mathrm{lora}\text{-}\mathrm{t}}_{Q},W^{\mathrm{lora}\text{-}\mathrm{t}}_{K},W^{\mathrm{lora}\text{-}\mathrm{t}}_{V},W^{\mathrm{lora}\text{-}\mathrm{t}}_O\right\}, \text{where}, \\
&W_{c}^{\mathrm{lora}\text{-}\mathrm{m}},W_c^{\mathrm{lora}\text{-}\mathrm{t}} = \mathrm{LoRA}_{\mathrm{MT}}\left(W_c\right),  c \in \left\{Q,K,V,O\right\},
\label{eq16}
\end{split}
\end{equation}
With these adaptations, the operations within a Transformer block equipped with modality-specific LoRA are defined as:
\begin{equation}
\begin{split}
\boldsymbol{h} &= \boldsymbol{x} + \mathrm{MHSA}\left(\mathrm{LN}\left(\boldsymbol{x}, \theta_{\mathrm{ln1}}\right); 
{\color{red} \theta_{\mathrm{attn}}^{\mathrm{lora}\text{-}\mathrm{u}}} \right), \ \mathrm{u} \in \left\{\mathrm{m},\mathrm{t}\right\}\\
\boldsymbol{x'} &= \boldsymbol{h} + \mathrm{FFN}\left(\mathrm{LN}\left(\boldsymbol{x}, \theta_{\mathrm{ln2}}\right); \theta_{\mathrm{ffn}}\right),
\label{eq17}
\end{split}
\end{equation}
where $\mathrm{MHSA}$ operates with modality-specific parameters  $\theta_{\mathrm{attn}}^{\mathrm{lora}\text{-}\mathrm{m}}$ or $\theta_{\mathrm{attn}}^{\mathrm{lora}\text{-}\mathrm{t}}$, as defined in Eq. \ref{eq16}, depending on whether the input is a motion or text token. In this design, the low-rank matrices $\Delta W^m$ and $\Delta W^t$ specialize in capturing modality-specific patterns, while the shared base matrix $W$ retains and refines the generalizable knowledge acquired during pre-training. This separation ensures efficient adaptation for heterogeneous modalities. 

\vspace{0.5\baselineskip} 

\noindent
\textbf{Mixture-of-Expert (MoE)}. \
In the MoE design, motion-specific parameters are designed as a set of specialized expert networks. These experts are seamlessly integrated into the LLM via the Mixture-of-Experts mechanism, which effectively mitigates modality interference between motion and language. 
To ensure stable training and fully leverage the pre-trained knowledge of the LLM, the motion experts are initialized using the weights of the Feed-Forward Networks (FFNs) from the pre-trained model. 

As shown in  Fig. \ref{fig2} (c), we introduce two distinct expert roles within the MoE framework: a \textbf{Motion-Expert}, tailored for processing motion-related features, and a \textbf{Text-Expert}, responsible for processing linguistic representations. To ensure effective modality separation, we adopt a \textit{static routing strategy}, which deterministically assigns motion and text tokens to their respective experts. This design ensures that motion-specific computations remain isolated from textual processing, thereby preserving the representational integrity of both modalities. Formally, the operations within a Transformer block augmented with MoE can be expressed as follows:
\begin{equation}
\begin{split}
\boldsymbol{h} &= \boldsymbol{x} + \mathrm{MHSA}\left(\mathrm{LN}\left(\boldsymbol{x}, \theta_{\mathrm{ln1}}\right); \theta_{\mathrm{attn}} \right), \\
\boldsymbol{x'} &= \boldsymbol{h} + \mathrm{FFN}\left(\mathrm{LN}\left(\boldsymbol{x}, \theta_{\mathrm{ln2}}\right); {\color{red} \theta^\mathrm{u}_{\mathrm{ffn}}}\right),  \ \mathrm{u} \in \left\{\mathrm{m},\mathrm{t}\right\}. 
\label{eq18}
\end{split}
\end{equation}
Here, $\mathrm{FFN}$ operates with modality-specific parameters $\theta^\mathrm{m}_{\mathrm{ffn}}$ and $\theta^\mathrm{t}_{\mathrm{ffn}}$, which are responsible for processing motion and text tokens, respectively.  Both sets of parameters are initialized using the corresponding FFN weights ($\theta_{\mathrm{ffn}}$ in Eq. \ref{eq12}) from the pretrained LLM. This design can effectively transfer valuable knowledge from the pretrained model while allowing each FFN to specialize in processing its respective modality. 

\vspace{0.5\baselineskip} 

\noindent
\textbf{Modality-Isolated Architecture (MIS)}.  \
In the MIS design, we enhance modular decoupling by incorporating modality-aware components. Specifically, we instantiate separate parameters for the Multi-Head Self-Attention ($\mathrm{MHSA}$), Layer Normalization ($\mathrm{LN}$), and Feed-Forward Network ($\mathrm{FFN}$) for each modality. This ensures that motion and text tokens are processed through entirely disjoint parameter pathways, allowing for modality-specific feature extraction while minimizing cross-modal interference. Under this architecture, the operations within a Transformer block can be formally defined as follows:
\begin{equation}
\begin{split}
\boldsymbol{h} &= \boldsymbol{x} + \mathrm{MHSA}\left(\mathrm{LN}\left(\boldsymbol{x}, {\color{red} \theta^{\mathrm{u}}_{\mathrm{ln1}}}\right); {\color{red} \theta^{\mathrm{u}}_{\mathrm{attn}}} \right), \\
\boldsymbol{x'} &= \boldsymbol{h} + \mathrm{FFN}\left(\mathrm{LN}\left(\boldsymbol{x}, {\color{red} \theta^{\mathrm{u}}_{\mathrm{ln2}}}\right); {\color{red} \theta^{\mathrm{u}}_{\mathrm{ffn}}}\right),   \ \mathrm{u} \in \left\{\mathrm{m},\mathrm{t}\right\}.
\label{eq19}
\end{split}
\end{equation}
Here, each layer operates with motion-related parameters $\left\{\theta^{\mathrm{m}}_{\mathrm{ln1}}, \theta^{\mathrm{m}}_{\mathrm{ln2}}, \theta^{\mathrm{m}}_{\mathrm{attn}}, \theta^\mathrm{m}_{\mathrm{ffn}}\right\}$ and text-related parameters $\left\{\theta^{\mathrm{t}}_{\mathrm{ln1}}, \theta^{\mathrm{t}}_{\mathrm{ln2}}, \theta^{\mathrm{t}}_{\mathrm{attn}}, \theta^\mathrm{t}_{\mathrm{ffn}}\right\}$. Motion and text tokens are processed independently through their respective parameter sets. 
All modality-specific parameters are initialized from the corresponding components of the pretrained LLM,  \ie, $\left\{\theta_{\mathrm{ln1}}, \theta_{\mathrm{ln2}},\theta_{\mathrm{attn}}, \theta_{\mathrm{ffn}} \right\}$ in Eq. \ref{eq12}. It allows the model to leverage prior knowledge while maintaining modality-specific specialization. 

\vspace{0.5\baselineskip} 

\noindent
\textbf{Parameter and Computational Cost Analysis}.  \
In this part, we first compare the parameter count of the four architecture variants. For clarity, we omit bias terms and layer normalization, as their impact is negligible compared to the dominant operations. Let $d_k$ denote the embedding dimension and $d_f$ the hidden size of the feed-forward network.  
The parameter count for a single Transformer block is then given by:
\begin{equation}
\begin{split}
&P_\mathrm{base} = 4d^2_k + 2d_fd_k \quad \text{(Prototype baseline)}\\
&P_{\mathrm{LoRA}}=4d^2_k + 2d_fd_k + {\color{blue} 16d_kr} \quad \text{(LoRA variant)}\\
&P_{\mathrm{MoE}}=4d_k^2 + {\color{blue} 4}d_fd_k \quad \text{(MoE variant)} \\
&P_{\mathrm{MIS}}={\color{blue} 2}\left(4d^2_k + 2d_fd_k\right) \quad \text{(MIS variant)}.
\end{split}
\end{equation}

Then, we compare the computational costs for four architectural variants.
Let $N$ be the number of input tokens.  The FLOPS for a multi-head self-attention (Eq. \ref{eq13}) are approximately $4N^2d_k+8Nd_k^2$, while the feed-forward layer (Eq. \ref{eq14}) requires about $4Nd_kd_{f}$. 
(1) \textbf{MOE}  and \textbf{MIS}: Both variants use a static routing strategy, where tokens are routed to modality-specific projections. As a result, they incur no extra computational cost. 
(2) \textbf{LoRA}: This variant injects low-rank matrices into the self-attention projections, adding an extra $16Nd_kr$ FLOPS (Eq. \ref{eq15}).
Accordingly, the total per-block FLOPs for each variant are summarized as follows:
\begin{equation}
\begin{split}
&F_{\mathrm{base}} = F_{\mathrm{MoE}} = F_{\mathrm{MIS}} = 4N^2d_k+8Nd_k^2+4Nd_kd_{f} \\
&F_{\mathrm{LoRA}} =4N^2d_k+8Nd_k^2+4Nd_kd_{f} + {\color{blue} 16Nd_kr}.
\end{split}
\end{equation}

To summarize, here is a comparison of the three architectural variants:
\begin{itemize}
\item LoRA: This variant adds only $16d_kr$ parameters and incurs an extra $16Nd_kr$ FLOPs through low-rank adapters in the MHSA layer. Since $ r\ll d_k$, the overhead is marginal, making LoRA highly parameter-efficient. 
\item MOE: This variant adds $2d_fd_k$  parameters, corresponding to a relative increase of $\frac{d_f}{2d_k+d_f}$ to the total parameter count, while introducing no additional FLOPs. This design effectively increases model capacity without a proportional rise in parameter number.
\item MIS: This variant doubles the total parameter count by separately instantiating all core components (MHSA, FFN, LN) for each modality. While it incurs no additional FLOPs due to parallel execution, it achieves maximal modality specialization at the expense of a two-fold increase in memory footprint. 
\end{itemize}

\subsection{Training Strategy}
\label{sec:3.4}
The training process consists of three stages. \textbf{Stage 1: Motion Tokenizer Training}: The initial stage focuses on training a motion tokenizer that learns to encode continuous motion data into multiple streams of discrete tokens. 
\textbf{Stage2: Modality-Alignment Pre-training}: The second stage aims to align motion and text modalities, enabling unified multimodal reasoning within the LLM. 
\textbf{Stage3: Multitask Instruction Fine-Tuning}: The third stage endows the model with comprehensive capabilities in motion comprehension, generation, reaction, and editing through supervised training across a diverse set of tasks, as detailed in Tab. \ref{tabA1}.

\vspace{0.5\baselineskip} 

\noindent
\textbf{Stage1: Motion Tokenizer Training.} \ In the first stage, the motion tokenizer is trained using the objective defined in Eq. \ref{eq4}. Once trained, it discretizes continuous 3D motion sequences into multiple parallel token streams. To capture both intra-stream and inter-stream dependencies, we apply the temporal delay mechanism (Eq. \ref{eq10}), which offsets token timings to facilitate cross-stream information flow. Notably, to ensure stability during subsequent LLM training, the motion tokenizer is frozen during the subsequent stages of training. 

\vspace{0.5\baselineskip} 

\noindent
\textbf{Stage2: Modality-Alignment Pre-training.} \ 
To enable MotionVerse to process discrete motion tokens effectively, we train it on a joint motion-text corpus. This training stage aims to align the motion and text modalities, facilitating unified reasoning within our parallel stream-aware LLM backbone. In this stage, we consider four basic tasks: two comprehension tasks (single-person and multi-person motion-to-text) and two generation tasks (single-person and multi-person text-to-motion). 
Formally, for a specific task, the source input is represented as a token sequence of length $T_s$, denoted by $\boldsymbol{q}_s=\left\{\boldsymbol{q}_s^i\right\}_{i=1}^{T_s}$, and the target output is a sequence of length $T_t$, denoted by $\boldsymbol{q}_t=\left\{\boldsymbol{q}_t^i\right\}_{i=1}^{T_t}$. Each token $\boldsymbol{q}_s^i$ or $\boldsymbol{q}_t^i$ corresponds either to a single text token or a group of $L$ motion tokens, the latter integrated using a temporal‐delay mechanism (Eq. \ref{eq10}).
The LLM models the conditional probability of each target token in an autoregressive manner: $p_{\theta}\left(\boldsymbol{q}_t^i|\boldsymbol{q}_t^{<i}, \boldsymbol{q}_s\right)$. Accordingly, the training objective is to maximize the log-likelihood of the observed token sequences:
\begin{equation}
\mathcal{L}\left(\theta\right)=\sum_{i=1}^{T_t}\log p_{\theta}\left(\boldsymbol{q}_t^i|\boldsymbol{q}_t^{<i},\boldsymbol{q}_s\right).
\end{equation}

\vspace{0.5\baselineskip} 

\noindent
\textbf{Stage3: Multitask Instruction Fine-Tuning.} \ 
After aligning the motion and text modalities, we adopt a multitask learning framework to address the tasks outlined in Tab. \ref{tabA1}.  A major challenge is \textbf{task interference}, which occurs when optimizing multiple objectives within a shared model leads to conflicting parameter updates. Specifically, the motion-to-text task, which generates natural language descriptions from 3D motion sequences, primarily relies on capturing \textit{high-level, global} motion features. In contrast, the motion-to-motion task (\eg, forecasting future motion frames) requires modeling  \textit{low-level, fine-grained} temporal dynamics. When these tasks are trained jointly, their divergent objectives can produce incompatible gradient directions, ultimately degrading the overall performance of the model. 

To mitigate task interference, we introduce a dedicated \textbf{task-specific motion tower} during the third stage of training. This tower is initialized with the parameters of the motion tower trained from Stage 2.  In this stage, motion tokens associated with motion-to-motion tasks (\eg, motion prediction) are routed through the specialized expert tower, while the original Stage 2 tower remains responsible for motion-to-text tasks.  This architectural separation allows each task type to operate with independent feature extractors, thereby maintaining task-specific representational capacity and substantially reducing conflicting gradient updates. Finally, we finetune the entire LLM model using the autoregressive objective over the full set of tasks listed in Tab. \ref{tabA1}. 

\section{Experiments}

\subsection{Datasets} \

We evaluate our approach on four publicly available datasets, MotionX \cite{lin2023motion}, InterHuman \cite{liang2024intergen}, InterX \cite{xu2024inter}, and MotionFix \cite{athanasiou2024motionfix}. 

\begin{table*}[t]
\centering
\caption{Quantitative evaluation results of text-to-motion (\textbf{T2M}) task on MotionX dataset.}
\begin{tabular}{l c c c c c c}
\toprule
\multirow{2}{*}{Methods} & \multicolumn{3}{c}{R-Precision (\%) $\uparrow$}  & \multirow{2}{*}{FID $\downarrow$} & \multirow{2}{*}{MM Dist $\downarrow$} & \multirow{2}{*}{Diversity $\rightarrow$ }\\
\cmidrule(lr){2-4}
& Top1  & Top2  & Top3  & & \\
\midrule
\color{mygray}{Real} & \color{mygray}{87.1} & \color{mygray}{93.9} & \color{mygray}{95.7} & - & \color{mygray}{0.696} & \color{mygray}{4.209} \\
MDM \cite{tevet2022human} & 71.3 & 82.8 & 86.2 & 0.034 & 0.872 & 4.166 \\
MotionDiffuse \cite{zhang2024motiondiffuse}  & 74.2 & 85.7 & 88.1 &  0.036 & 0.852 & 3.898   \\
T2M-GPT \cite{zhang2023generating} & 74.7 & 86.1& 88.5& 0.038 & 0.844 & 4.339 \\ 
MMM \cite{pinyoanuntapong2024mmm} & 77.2 & 87.4 & 90.5 & 0.017 & 0.817 & 4.141 \\ 
MoMask \cite{guo2024momask} & \textbf{78.5} & 87.9 & 91.1 & \textbf{0.007} & 0.835 & 4.322 \\ 
\midrule
MotionGPT-13B \cite{zhang2024motiongpt} & 54.8 & 72.7 & 79.3 & 0.076 & 0.907 & 4.025 \\ 
MotionGPT \cite{jiang2023motiongpt} & 75.1 & 86.4 & 88.7 & 0.041 & 0.833 & 4.359 \\
M$^3$GPT \cite{luo2024m} & 75.3 & 86.5 & 88.9 & 0.038 & 0.830 & 4.363\\
%\midrule
{\color{blue} MotionVerse (ours)} & \textbf{78.5} & \textbf{88.2} & \textbf{91.3} & 0.010 &\textbf{0.811} & \textbf{4.222}   \\
\bottomrule
\end{tabular}
\label{tab1}
\end{table*}
      
\begin{table*}[t]
\centering
\caption{Quantitative evaluation results of the interactive text-to-motion (\textbf{I-T2M}) task on InterHuman and InterX dataset.}
\begin{tabular}{l l c c c c c c}
\toprule
\multirow{2}{*}{Dataset} & \multirow{2}{*}{Methods} & \multicolumn{3}{c}{R-Precision (\%) $\uparrow$}  & \multirow{2}{*}{FID $\downarrow$} & \multirow{2}{*}{MM Dist $\downarrow$} & \multirow{2}{*}{Diversity $\rightarrow$ }\\
\cmidrule(lr){3-5}
& & Top1 & Top2 & Top3 & & \\
\midrule
\multirow{4}{*}{InterHuman}
&\color{mygray}{Real} & \color{mygray}{52.3} & \color{mygray}{67.4} & \color{mygray}{74.0} &-& \color{mygray}{1.042} & \color{mygray}{4.199}\\
&ComMDM \cite{shafir2023human} & 34.1 & 46.2 & 60.3 & 0.242 &1.723 & 3.827 \\ 
&InterGen \cite{liang2024intergen} & 47.6 & 58.8 & 71.4 & 0.203 & 1.417 & 3.903\\
&{\color{blue} MotionVerse (ours)} & \textbf{56.3} & \textbf{70.6} & \textbf{76.9} & \textbf{0.064} & \textbf{1.012} & \textbf{4.164} \\
\midrule
\multirow{4}{*}{InterX}
&\color{mygray}{Real} & \color{mygray}{76.5} & \color{mygray}{89.5} & \color{mygray}{93.9} & -& \color{mygray}{0.819} & \color{mygray}{4.226}\\
&ComMDM \cite{shafir2023human} &  59.4 & 73.0 & 82.2 & 0.108 & 1.355 & 3.852 \\ 
&InterGen \cite{liang2024intergen} & 68.2 & 81.5 & 89.0 & 0.091 & 1.066 & 4.021 \\
&{\color{blue} MotionVerse (ours)} & \textbf{71.9} & \textbf{87.1} & \textbf{91.7} & \textbf{0.023} & \textbf{0.838} & \textbf{4.198} \\
\bottomrule
\end{tabular}
\label{tab2}
\end{table*}

\noindent
\textbf{MotionX.} \  MotionX \cite{lin2023motion} is a large-scale 3D human motion dataset comprising 95,641  motion sequences, each paired with corresponding textural descriptions. It is constructed from approximately  15,000 monocular videos collected from various online platforms and publicly available video datasets. MotionX captures a wide range of human behaviors, spanning everyday activities, social interactions, and sports movements, making it highly diverse in both motion types and environmental contexts.

\vspace{0.5\baselineskip} 
\noindent
\textbf{InterHuman and InterX.} InterHuman \cite{liang2024intergen} and InterX \cite{xu2024inter} are two large-scale datasets focused on human-human interactions, each accompanied by textual annotations. 
\textbf{InterHuman} \cite{liang2024intergen} covers a broad spectrum of two-person interactions, from everyday activities (such as hugging) to specialized motions (such as boxing and dancing). It consists of 6,022 motion sequences with 1.7 million frames. 
\textbf{InterX} \cite{xu2024inter}  extends this effort by employing an advanced motion capture system based on optical technology to precisely record body movements. It comprises 11,388 motion sequences across 40 categories of daily interactions, amounting to over 8.1 million frames.

\vspace{0.5\baselineskip} 
\noindent
\textbf{MotionFix.}  MotionFix \cite{athanasiou2024motionfix} is the first dataset designed for language-driven motion editing, allowing fine-grained manipulation of 3D human motions via natural language instructions.  
It supports open-ended and diverse editing operations, including spatial edit (\eg, ``\textit{throw from higher}"), temporal edit  (\eg, ``\textit{start standing not bent down}") and composite edits (\eg, ``\textit{bend down a bit more, stand up faster"}").  The dataset comprises 4,771 triplets, each consisting of a source motion, a target motion, and a corresponding textual description that specifies the desired modification. 

\vspace{0.5\baselineskip} 

%\noindent
\subsection{Evaluation Metrics} 
Different tasks employ distinct evaluation metrics. Following \cite{jiang2023motiongpt,zhang2024motiongpt}, we adopt widely-used metrics specific to each task to assess the performance of our method. 

\vspace{0.5\baselineskip} 
\noindent
 \textbf{Text-to-Motion.} \ For the text-to-motion task involving single-person and multi-person interactions, we employ the following evaluation metrics: (1) \textit{Frechet Inception Distance} (FID), which measures the distributional difference between high-level features of generated and real motions, providing an assessment of overall motion quality; (2) \textit{R-Precision} and \textit{MultiModal Distance} (MM Dist), which evaluate semantic alignment between input texts and generated motions. Notably, R-precision is calculated using batches of $32$ samples. (3) \textit{Diversity} (Div), which measures the variability among motions generated from the same textual prompt. 

\vspace{0.5\baselineskip} 
\noindent 
\textbf{Motion-to-Text.} \ For the motion-to-text task, including single-person and multi-person motion captioning, we adopt the following evaluation metrics: (1) \textit{R-Precision} and \textit{MultiModal Distance} (MM Dist), which measure semantic alignment between generated texts and corresponding ground-truth motions; 
(2) \textit{Linguistic metrics}, including BLEU, ROUGE, and BERTScore, which respectively assess the quality of the generated descriptions through n-gram overlap, recall-oriented summarization, and contextual embedding similarity. 

\vspace{0.5\baselineskip} 
\noindent
\textbf{Motion-to-Motion.} \ For the motion-to-motion task, including motion prediction and motion in-between tasks, we employ the following evaluation metrics: (1) \textit{FID} and \textit{Diversity} (Div) to evaluate motion quality and variability; (2)  \textit{Average Displacement Error} (ADE) and \textit{Final Displacement Error} (FDE) to measure the accuracy of predicted motions related to ground-truth sequences. 

\vspace{0.5\baselineskip} 
\noindent
\textbf{Motion+Text-to-Motion.} \ For the motion+text-to-motion task, including motion reaction and motion editing, we use the following evaluation metrics: (1) \textit{FID} and \textit{Diversity} (Div) to assess the quality and variability of generated motions; (2) \textit{R-Precision} and \textit{MultiModal Distance} (MM Dist) to measure semantic alignment between the combined motions (integrating both input and generated motions) and their corresponding textual descriptions. 

\vspace{0.5\baselineskip} 
\noindent
\textbf{Retrieval Networks.} \ 
To calculate the above metrics, we train three specialized retrieval networks tailored to distinct motion-text alignment tasks: 
(1) \textbf{Single-person motion retrieval}:  This model consists of a single-person motion encoder implemented using six transformer \cite{dosovitskiy2020image} layers and a BERT-based \cite{devlin2019bert} text encoder. It employs contrastive learning to optimize the joint embedding space by minimizing distances between matched motion-text pairs and maximizing separation for negative samples.
(2) \textbf{Multi-person interaction retrieval}: We construct an interaction-aware retrieval model, comprising an interaction motion encoder (six transformer layers \cite{dosovitskiy2020image})  paired with a  BERT-based \cite{devlin2019bert} text encoder. Similarly, contrastive learning is applied to obtain discriminative embeddings specifically for interaction motion-text pairs.
(3) \textbf{Motion pair retrieval}: This model comprises a motion-pair encoder with six transformer layers \cite{dosovitskiy2020image} to process pre- and post-edit motion sequences, coupled with a CLIP-based \cite{radford2021learning} text encoder designed to encode textural edit descriptions. A contrastive loss is used to align motion pairs with their corresponding textual edits within a shared embedding space. 

\begin{table*}[t]
\centering
\caption{Quantitative evaluation results of motion-to-text (\textbf{M2T}) task on MotionX, and interactive motion-to-text (\textbf{I-M2T}) task on InterHuman and InterX datasets.}
\begin{tabular}{l l c c c c c c c c }
\toprule
\multirow{2}{*}{Dataset} & \multirow{2}{*}{Methods} & \multicolumn{3}{c}{R-Precision (\%) $\uparrow$}  & \multirow{2}{*}{MMDist $\downarrow$} & \multirow{2}{*}{BLEU@1 $\uparrow$} &\multirow{2}{*}{BLEU@4 $\uparrow$} & \multirow{2}{*}{ROUGE-L $\uparrow$ } & \multirow{2}{*}{BERT Score $\uparrow$ } \\
\cmidrule(lr){3-5}
& & Top1 & Top2 & Top3 & & & & \\
\midrule
\multirow{3}{*}{MotionX}
&\color{mygray}{Real} &\color{mygray}{87.1} & \color{mygray}{93.9} & \color{mygray}{95.7} & \color{mygray}{0.696} & - &-&-&- \\
&TM2T \cite{guo2022tm2t} & 74.6 & 83.5 & 86.1 & 0.834 & 36.3 & 8.1 & 38.4 & 33.2 \\
&MotionGPT \cite{jiang2023motiongpt} & 83.1 & 88.0 & 89.7 & 0.750 & 34.2 & 9.5 & 36.0 & 33.9  \\
&M$^3$GPT \cite{luo2024m} & 83.5 & 88.3 & 90.1 & 0.753 & 35.5 & 12.2 & 35.6 & 34.3 \\
&{\color{blue} MotionVerse (ours)} & \textbf{84.2} &\textbf{90.8} &\textbf{92.9} &\textbf{0.702} & \textbf{39.6} & \textbf{15.8} &\textbf{38.8} &\textbf{38.8} \\
\midrule
\multirow{3}{*}{InterHuman}
&\color{mygray}{Real} & \color{mygray}{52.3} & \color{mygray}{67.4} & \color{mygray}{74.0} & \color{mygray}{1.042} & - &- &- &- \\
&TM2T \cite{guo2022tm2t} & 45.3 & 58.5 & 65.7 & 1.233 & 28.9 & 6.6 & 22.1 & 18.5 \\
&{\color{blue} MotionVerse (ours)} & \textbf{49.9} & \textbf{62.9} & \textbf{69.1} & \textbf{1.036} & \textbf{34.1} &\textbf{8.0}&\textbf{25.9} &\textbf{20.4}   \\
\midrule
\multirow{3}{*}{InterX}
&\color{mygray}{Real} &\color{mygray}{76.5} & \color{mygray}{89.5} & \color{mygray}{93.9} & \color{mygray}{0.819} & - &- &- &- \\
&TM2T \cite{guo2022tm2t} & 66.0 & 78.7 & 83.4 & 0.985 & 40.3 & 16.8 & 36.6 & 29.1 \\
&{\color{blue} MotionVerse (ours)} &\textbf{70.2} & \textbf{83.6} & \textbf{88.7} &\textbf{0.862} &\textbf{44.1} &\textbf{18.5} & \textbf{39.1} & \textbf{33.8}  \\
\bottomrule
\end{tabular}
\label{tab3}
\end{table*}

\subsection{Implementation Details} 

Following \cite{liang2024intergen}, we adopt a non-canonical motion representation to effectively model multi-person interactions. This representation is a $263$-dimensional vector comprising global joint positions and velocities  (in the world coordinate frame), local joint rotations (relative to the root joint frames), and binary foot-ground contact indicators.  

\vspace{0.5\baselineskip} 
\noindent
\textbf{Motion Tokenizer Training.} \ 
To train the motion tokenizer, we implement a residual VQ-VAE architecture with a downscaling factor of $4\times$, employing residual blocks in both encoder and decoder modules. The tokenizer consists of $L=6$ quantization layers, each with a codebook of $K=512$ codes. During training, input motion sequences are cropped to a fixed length of $64$ frames. The model is optimized using AdamW \cite{loshchilov2017decoupled} with a batch size of $128$ over $100$K iterations. We apply a linear warm-up schedule to the learning rate, reaching $2\times10^{-4}$ after the first $1000$ iterations. The loss weighting parameter  
$\beta$ in the residual VQ loss ($\mathcal{L}_{\mathrm{rvq}}$, Eq. \ref{eq4}) is set to $0.02$.

\vspace{0.5\baselineskip} 
\noindent
\textbf{MotionVerse Framework Training.} \ 
In the MotionVerse framework, we adopt T5-base \cite{raffel2020exploring} as the backbone language model. We employ AdamW optimizer \cite{loshchilov2017decoupled} along with a CosineAnealingLR learning rate scheduler. During the pre-training stage, the model is trained  $400$K iterations with an initial learning rate of $1 \times 10^{-3}$. In the subsequent multitask fine-tuning stage, training continues for another $400$K iterations with a reduced initial learning rate of  $5 \times 10^{-4}$. For the Motion-Fusion LoRA variant (Eq. \ref{eq15}),  we set the rank $r$ to $8$ and the scaling factor $\alpha$  to $16$.
All experiments are conducted on 2 NVIDIA A100 GPUs.

\subsection{Comparison with State-of-the-arts}
In this section, we compare our MotionVerse with state-of-the-art (SOTA) methods across multiple core motion-relevant tasks listed in Tab. \ref{tabA1}. To enable fair and consistent comparison, we replicated prior SOTA approaches. \textbf{Importantly, MotionVerse adopts a unified multi-task architecture, utilizing a shared model structure and parameter set across all tasks.}

\vspace{0.5\baselineskip} 
\noindent
\textbf{Comparisons on Text-to-Motion.} \
Tab. \ref{tab1} summarizes the comparison results on Text-to-Motion (T2M) task. We can observe that: (1) Compared to task-specific methods \cite{tevet2022human,zhang2024motiondiffuse,zhang2023generating,pinyoanuntapong2024mmm}, MotionVerse achieves superior performance on most metrics. Although our method yields a slightly higher FID score compared to MoMask \cite{guo2024momask}, this is primarily due to MoMask's specialized architecture tailored for T2M generation. Specifically, MoMask utilizes two separate generative networks for base and residual motion tokens, effectively increasing model capacity but sacrificing flexibility. In contrast, MotionVerse demonstrates significantly broader generalizability, seamlessly integrating up to ten distinct motion tasks within a unified framework while maintaining competitive performance across all tasks.
(2) Existing multi-task frameworks \cite{zhang2024motiongpt,jiang2023motiongpt,luo2024m} typically exhibit inferior performance compared to task-specific methods. In contrast, MotionVerse not only outperforms these multi-task approaches but also matches or surpasses the performance of dedicated single-task models. These results demonstrate the effectiveness of our framework. 

\begin{figure*}
\centering
\captionsetup{font={small}}
\includegraphics[width=0.8\linewidth]{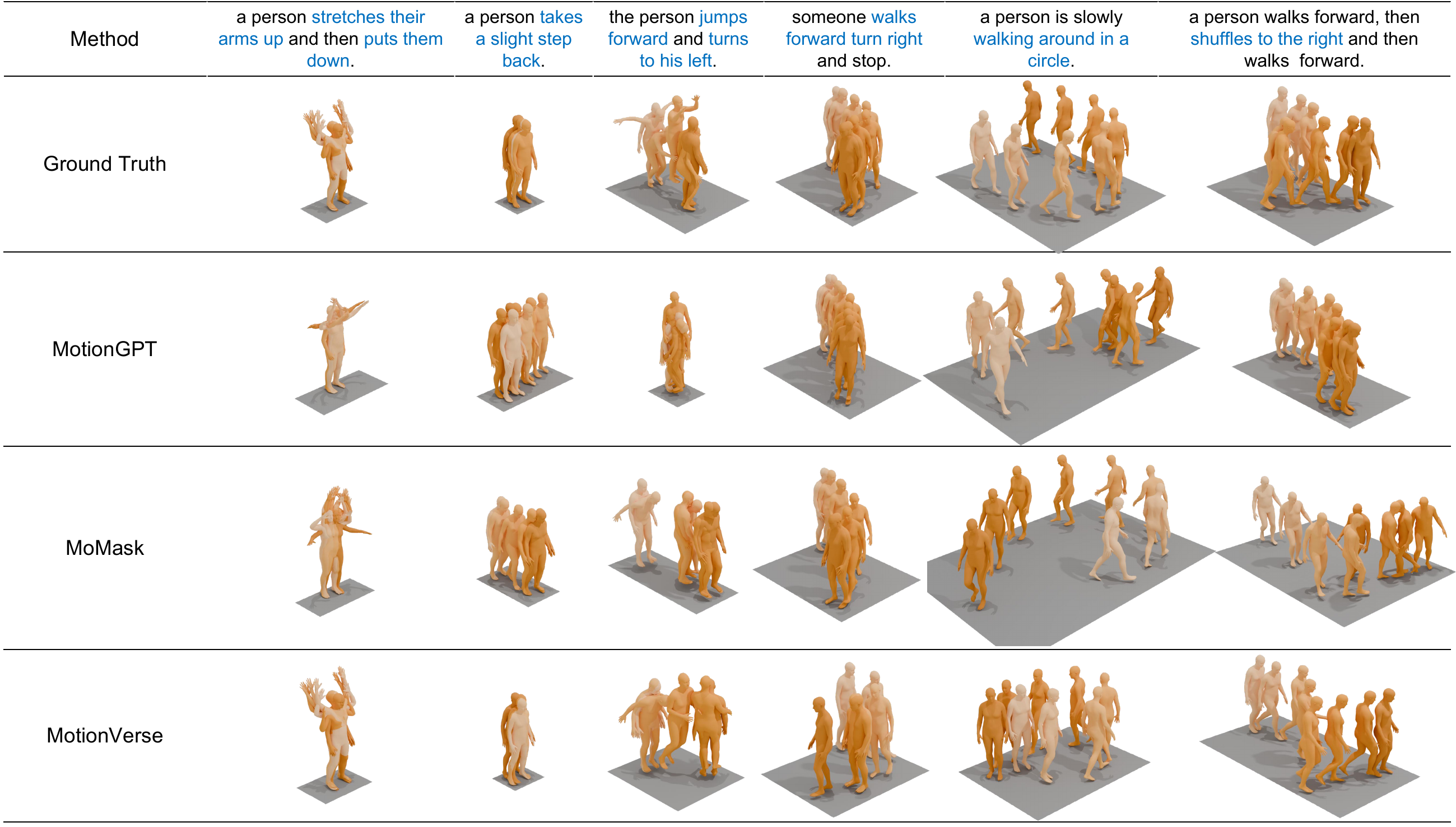}
\caption{Qualitative evaluation on text-to-motion generation. We qualitatively compared the visualizations generated by our method with those produced by MotionGPT \cite{jiang2023motiongpt}  and MoMask \cite{guo2024momask}.}
\label{fig5}
\vspace{-5pt}
\end{figure*}
\begin{figure*}
\centering
\captionsetup{font={small}}
\includegraphics[width=0.8\linewidth]{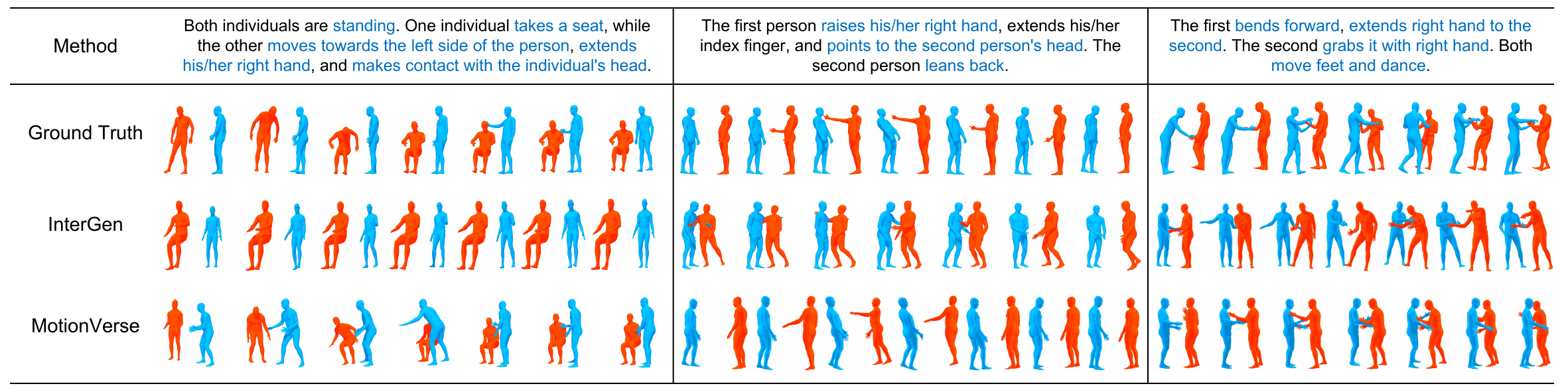}
\caption{Qualitative evaluation on interactive text-to-motion generation. We qualitatively compared the visualizations generated by our method with those produced by InterGen \cite{liang2024intergen}.}
\label{fig6}
\vspace{-10pt}
\end{figure*}

Fig. \ref{fig5} presents the visualization results for the text-to-motion generation task. We compare our method with MotionGPT \cite{jiang2023motiongpt} and MoMask \cite{guo2024momask}. It can be observed that the motions generated by MotionGPT often align with only the initial segment of the textual description while ignoring the latter part. In contrast, our method yields motions that demonstrate a more faithful alignment with the entire text, thereby capturing the complete semantic content of the description. These visualization results further validate the effectiveness of our approach in generating semantically consistent and coherent human motions. 

\vspace{0.5\baselineskip} 

\noindent
\textbf{Comparisons on Interactive Text-to-Motion.} \
Tab. \ref{tab2} summarizes the comparison results for the Interactive Text-to-Motion (I-T2M) task. MotionVerse consistently outperforms other baselines \cite{shafir2023human,liang2024intergen} across all evaluation metrics. These results highlight the superior capability of our framework in generating more realistic interactive motions while maintaining precise alignment with input textual descriptions.  The results demonstrate that  MotionVerse excels not only in single-person scenarios but also in more complex two-person interactive settings, underscoring its robustness in modeling both individual and social motion dynamics.

Fig. \ref{fig6} presents the visualization results for the interactive text-to-motion generation task. We compare our method with InterGen \cite{liang2024intergen}. The visualization results demonstrate that our method is capable of generating more realistic two-person interactions while exhibiting stronger adherence to the semantics of the input text.

\begin{table*}[t]
\centering
\caption{Quantitative evaluation results of Motion Prediction and In-between tasks on MotionX (single-person), InterHuman and InterX (interactive two-person) datasets.}
\begin{tabular}{l l c c c c c c c c c }
\toprule
\multirow{2}{*}{Dataset} & \multirow{2}{*}{Methods} & \multicolumn{4}{c}{Motion Prediction}  & \multicolumn{4}{c}{Motion In-between} \\
\cmidrule(lr){3-6}  \cmidrule(lr){7-10} 
& & FID $\downarrow$ & Div $\rightarrow$ & ADE $\downarrow$ & FDE $\downarrow$ & FID $\downarrow$ & Div $\rightarrow$ & ADE $\downarrow$ & FDE $\downarrow$  \\
\midrule
\multirow{6}{*}{MotionX}
&\color{mygray}{Real} & - & \color{mygray}{4.213} & - &- & - & \color{mygray}{4.194} & - & - \\ 
&MDM \cite{tevet2022human} & 0.046 & 4.298 & 2.430 & 3.871 & 0.044 & 4.283 & 1.995 & 2.862 \\
&MotionGPT-13B \cite{zhang2024motiongpt} & 0.082 & 4.174 & 2.634 & 3.825 & 0.079 & 4.131 & 2.036 & 2.910  \\
&MotionGPT \cite{jiang2023motiongpt} & 0.047 & 4.279 & 2.412 & 3.558 & 0.039 & 4.240 & 1.931 & 2.828 \\ 
&M$^3$GPT   \cite{luo2024m} & 0.043 & 4.357 & 2.399 & 3.516 & 0.032 & 4.349 & 1.908 & 2.853   \\
&{\color{blue} MotionVerse (ours)} & \textbf{0.011} & \textbf{4.232} &\textbf{2.136} &\textbf{3.303} &\textbf{0.007} & \textbf{4.224} &\textbf{1.780} & \textbf{2.776} \\ 
\midrule
\multirow{3}{*}{InterHuman}
& \color{mygray}{Real}  & - & \color{mygray}{4.167} &- &- & - & \color{mygray}{4.187} &- &- \\
&MDM \cite{tevet2022human} & 0.089 & 4.283 & 1.766 & 2.026 & 0.083 & 4.262 & 1.613 & 1.825 \\
&{\color{blue} MotionVerse (ours)} & \textbf{0.042} & \textbf{4.120} & \textbf{1.329} & \textbf{1.821} & \textbf{0.037} &\textbf{4.173} & \textbf{1.160} & \textbf{1.583}   \\ 
\midrule
\multirow{3}{*}{InterX}
& \color{mygray}{Real} & - &\color{mygray}{4.222} & - &- & - & \color{mygray}{4.218} &- & -\\ 
&MDM \cite{tevet2022human} & 0.075 & 4.320 & 1.206 & 1.307 & 0.071 & 4.282 & 1.058 & 1.206  \\
&{\color{blue} MotionVerse (ours)}& \textbf{0.036} & \textbf{4.194} & \textbf{0.914} & \textbf{1.079} &\textbf{0.031} & \textbf{4.204} & \textbf{0.837} & \textbf{0.987} \\ 
\bottomrule
\end{tabular}
\label{tab4}
\end{table*}
      
\begin{table*}[t]
\centering
\caption{Quantitative evaluation results of the motion reaction (\textbf{React}) task on InterHuman and InterX datasets.}
\begin{tabular}{l l c c c c c c}
\toprule
\multirow{2}{*}{Dataset} & \multirow{2}{*}{Methods} & \multicolumn{3}{c}{R-Precision (\%) $\uparrow$}  & \multirow{2}{*}{FID $\downarrow$} & \multirow{2}{*}{MM Dist $\downarrow$} & \multirow{2}{*}{Diversity $\rightarrow$ }\\
\cmidrule(lr){3-5}
& & Top1 & Top2 & Top3 & & \\
\midrule
\multirow{4}{*}{InterHuman}
&\color{mygray}{Real} & \color{mygray}{52.3} & \color{mygray}{67.4} & \color{mygray}{74.0} & - & \color{mygray}{1.042} & \color{mygray}{4.199} \\
&MDM \cite{tevet2022human} & 37.2 & 52.5 & 58.4 & 1.073 & 1.518 & 4.330 \\ 
&ReGenNet \cite{xu2024regennet} & 48.8 & 63.1 & 71.5 & 0.063 & 1.124 & 4.210 \\
&{\color{blue} MotionVerse (ours)} &\textbf{50.3} & \textbf{65.3} & \textbf{72.8} & \textbf{0.042} & \textbf{1.052} & \textbf{4.168}  \\
\midrule
\multirow{4}{*}{InterX}
&\color{mygray}{Real} & \color{mygray}{76.5} & \color{mygray}{89.5} & \color{mygray}{93.9} & -& \color{mygray}{0.819} & \color{mygray}{4.226} \\
&MDM \cite{tevet2022human} & 66.3 & 80.8  & 85.1 & 0.087 & 1.430 & 4.331 \\ 
&ReGenNet \cite{xu2024regennet} & 70.8 & 85.3 & 89.6 & 0.051 & 0.905 & 4.157 \\
&{\color{blue} MotionVerse (ours)} & \textbf{71.6} & \textbf{85.7} & \textbf{90.7} &\textbf{0.017} & \textbf{0.863} & \textbf{4.189} \\
\bottomrule
\end{tabular}
\label{tab5}
\end{table*}
      
\begin{table*}[t]
\centering
\caption{Quantitative evaluation results of the motion editing (\textbf{Edit}) task on MotionFix dataset.}
\begin{tabular}{l c c c c c c}
\toprule
 \multirow{2}{*}{Methods} & \multicolumn{3}{c}{R-Precision (\%) $\uparrow$}  & \multirow{2}{*}{FID $\downarrow$} & \multirow{2}{*}{MM Dist $\downarrow$} & \multirow{2}{*}{Diversity $\rightarrow$ }\\
\cmidrule(lr){2-4}
& Top1 & Top2 & Top3 & & \\
\midrule
\color{mygray}{Real} & \color{mygray}{46.1} & \color{mygray}{63.3} & \color{mygray}{70.6} & - & \color{mygray}{1.182} & \color{mygray}{3.941} \\
MDM \cite{tevet2022human} & 29.2 & 38.7 & 45.3 & 1.104 & 1.877 & 4.226\\ 
TMED \cite{athanasiou2024motionfix} & 35.8 & 50.4 & 55.9 & 0.061 & 1.503 & 3.740\\
{\color{blue} MotionVerse (ours)} & \textbf{38.7} & \textbf{54.8} & \textbf{62.3} & \textbf{0.029} & \textbf{1.209} & \textbf{3.968}   \\
\bottomrule
\end{tabular}
\label{tab6}
\end{table*}

\vspace{0.5\baselineskip} 
\noindent
\textbf{Comparisons on  Motion-to-Text.} \ 
Tab. \ref{tab3} summarizes the comparison results for Motion-to-Text (M2T) and Interactive  Motion-to-Text (I-M2T) tasks. Our generalist model, MotionVerse,  achieves the best performance across all evaluation metrics on both tasks, significantly surpassing previous methods including TM2T \cite{guo2022tm2t}, MotionGPT \cite{jiang2023motiongpt} and M$^3$GPT \cite{luo2024m}. 
The performance gain stems from MotionVerse's integration of LLMs and decoupled motion-text modeling, which enables effective incorporation of advanced linguistic knowledge. The improvements manifest in two key aspects: (1) Enhanced text quality, as evidenced by significant gains in standard linguistic metrics (BLEU, ROUGE, and BERTScore), indicating superior grammatical fluency and logical coherence; and (2) Improved motion-text alignment, demonstrated by higher Top-1 Retrieval Accuracy and lower MM-Dist scores. These quantitative results collectively confirm that MotionVerse generates descriptions that exhibit both high linguistic quality and precise semantic correspondence with the input motion.

\vspace{0.5\baselineskip} 
\noindent
\textbf{Comparisons on Motion Prediction and In-between.} \ 
Tab. \ref{tab4} presents the comparison results for the Motion Prediction and In-between tasks across three benchmark datasets: MotionX, InterHuman, and InterX. In the Motion Prediction task, models are conditioned on only the initial $20\%$ of the motion sequence and tasked with generating a plausible future trajectory. In the Motion In-between task, approximately $50\%$ of the sequence is masked, requiring the model to generate realistic and coherent motions to fill the missing segment. As shown in Tab. \ref{tab4}, MotionVerse notably outperforms other approaches, including MDM \cite{tevet2022human}, MotionGPT \cite{ zhang2024motiongpt, jiang2023motiongpt}, M$^3$GPT \cite{ luo2024m}, across FID, diversity, and displacement error. This superior performance highlights our model's ability to generate high-quality, diverse motions while minimizing kinematic errors.

\vspace{0.5\baselineskip} 
\noindent
\textbf{Comparisons on Motion Reaction.} \
Tab. \ref{tab5} shows the comparisons on the Motion Reaction task. MotionVerse 
 achieves significantly higher Top-1 Retrieval Accuracy, lower FID, and reduced Multi-Modal Distance compared to existing methods, including MDM \cite{tevet2022human} and ReGenNet \cite{xu2024regennet}. These metrics collectively highlight MotionVerse's superior capability in generating realistic, temporally coherent, and semantically appropriate reactions in interactive motion scenarios. We attribute this performance advantage to MotionVerse's autoregressive modeling coupled with multi-level motion tokenization, which enable fine-grained temporal conditioning and stronger semantic coupling between interacting agents.

\vspace{0.5\baselineskip} 
\noindent
\textbf{Comparisons on Motion Editing.} \
Tab. \ref{tab6} shows the comparisons on the Motion Editing task. Our MotionVerse clearly outperforms the existing approaches, including MDM \cite{tevet2022human} and TMED \cite{athanasiou2024motionfix}, acorss all evaluation metrices.  This superior performance demonstrates the effectiveness of our unified framework in motion editing.   

\vspace{0.5\baselineskip}

% \noindent
% \textbf{Comparisons on Motion Diff.} 

\subsection{Ablation Studies}
This section presents a comprehensive ablation study to evaluate the key design components of MotionVerse. 
Unless otherwise noted, the default configuration uses \textbf{Delay Parallel} and \textbf{Motion-Fusion with Mixture-of-Experts (MoE)} as the foundational components.

\begin{table}[t]
\centering
\caption{Reconstruction and Generation Performance of Residual Vector Quantization (RVQ) vs. Vanilla Vector Quantization (VQ) on MotionX. We evaluate the Generation Performance on the single-task Text-to-Motion (T2M).  $L$ denotes the number of quantization layers in RVQ, and  MPJPE is reported in millimeters.}
\begin{tabular}{l l c c c c}
\toprule
\multirow{2}{*}{Methods} & \multirow{2}{*}{$L$}  & \multicolumn{2}{c}{Reconstruction} & \multicolumn{2}{c}{Generation (T2M)} \\
\cmidrule(lr){3-4} \cmidrule(lr){5-6}
&  & FID $\downarrow$ & MPGPE $\downarrow$ & FID $\downarrow$ & MM-Dist  $\downarrow$ \\ 
\midrule
VQ & - &0.023 &72.8 & 0.030 & 0.894 \\
\midrule
RVQ & 2 & 0.011 & 58.0 &0.024 &0.854 \\
RVQ & 4 &0.004 & 46.7 &0.015 & 0.802 \\
RVQ & 6  & 0.002 & 41.7 & \textbf{0.012} & \textbf{0.791} \\
RVQ & 8 & \textbf{0.001} &\textbf{37.2} & 0.013 & 0.795\\
\bottomrule
\end{tabular}
\label{tab_A1}
\end{table}
\begin{table}[t]
\centering
\caption{Ablation study on multi-stream motion token modeling strategies, evaluating all models after training on the  \textbf{single Text-to-Motion} task  on MotionX.}
\begin{tabular}{l c c c}
\toprule
Models & Top1 (\%) $\uparrow$  & FID $\downarrow$ & MM Dist $\downarrow$ \\
\midrule
Single-stream (VQ) & 69.5 &0.030 &0.894 \\
\midrule
Flattening & 78.8 & \textbf{0.012} & 0.802\\
Parallel & 64.7 & 0.036 & 0.958\\
Delay Parallel & \textbf{79.5} & \textbf{0.012} & \textbf{0.791} \\
\bottomrule
\end{tabular}
\label{tab_A2}
\end{table}

\vspace{0.5\baselineskip} 
\noindent
\textbf{The effect of Residual Quantization.} \ 
Tab. \ref{tab_A1} provides a detailed comparison between Residual Vector Quantization (RVQ) and standard Vector Quantization (VQ) in terms of both \textbf{reconstruction} and \textbf{generation} performance on the MotionX dataset. As shown in Tab. \ref{tab_A1}, RVQ consistently outperforms VQ in motion reconstruction, achieving lower FID and MPGPE scores,  which indicates higher fidelity and structural accuracy of the reconstructed motions. Furthermore, we train our model on the single-task Text-to-Motion setting using RVQ and VQ tokens, respectively.  RVQ yields more faithful motion generation, enabling our model to generate more coherent and semantically aligned motions with the input conditions. These results demonstrate the effectiveness of RVQ in capturing richer motion representations and improving the overall quality of motion generation.

\vspace{0.5\baselineskip} 
\noindent
\textbf{Number of Residual Layers ($L$).} \ In Tab. \ref{tab_A1}, we investigate RVQ with different numbers of quantization layers ($L$). 
Generally, as $L$ increases, the reconstruction error decreases steadily, indicating a more precise approximation of the original motion sequence. However, increasing $L$ also expands the token space, thereby imposing a greater modeling burden on the language model during generation.  In Tab.  \ref{tab_A1}, we observe a drop in generation quality when $L=8$, likely due to the increased difficulty of accurately predicting more residual tokens. 
Based on these empirical observations, we fix the number of residual layers to $L=6$ in our implementation. This choice strikes a balance between the expressive power of RVQ and the quality of generation performance. 

\begin{table}[t]
\centering
\caption{Ablation study on architectural variants for modality-specific towers in Motion-Fusion. The \textbf{Single-Task} adopts the Prototype architecture and is trained solely on the \textbf{Text-to-Motion (T2M)} task.  Other models are trained on all tasks. All models are evaluated on T2M task.}
\begin{tabular}{l c c c c c c}
\toprule
\multirow{2}{*}{Models} & \multicolumn{2}{c}{MotionX} & \multicolumn{2}{c}{InterHuman} & \multicolumn{2}{c}{InterX} \\
\cmidrule(lr){2-3} \cmidrule(lr){4-5} \cmidrule(lr){6-7}
& Top1 $\uparrow$  & FID $\downarrow$ & Top1 $\uparrow$  & FID $\downarrow$  & Top1 $\uparrow$  & FID $\downarrow$  \\
\midrule
Single-Task  & 80.7 & 0.009 & 54.4 & 0.072 & 73.3 & 0.024 \\
\midrule
Protoype &74.9 &0.012 &50.7 & 0.076 &63.8 & 0.027\\
LoRA & 77.1 & 0.011 &53.3 & 0.069 & 70.2 & 0.024\\
MoE & 78.5 & \textbf{0.010} & \textbf{56.3} & \textbf{0.064} & 71.9 & \textbf{0.023}\\
MIS  & \textbf{79.1} & \textbf{0.010} & 54.9 & 0.067 & \textbf{72.4} & \textbf{0.023}\\ 
\bottomrule
\end{tabular}
\label{tab_A4}
\end{table}
\begin{table*}[t]
\centering
\caption{Ablation study of the Task-Level Motion Tower (TMT) and Large Language Backbone on Text-to-Motion and Motion-to-Text tasks.}
\begin{tabular}{l c c c c c c c c c c c c  }
\toprule
\multirow{2}{*}{Models} & \multicolumn{6}{c}{Text-to-Motion} & \multicolumn{6}{c}{Motion-to-Text}  \\
\cmidrule(lr){2-7}  \cmidrule(lr){8-13}   
&\multicolumn{2}{c}{MotionX} &\multicolumn{2}{c}{InterHuman} &\multicolumn{2}{c}{InterX} &\multicolumn{2}{c}{MotionX} &\multicolumn{2}{c}{InterHuman} &\multicolumn{2}{c}{InterX}  \\
\cmidrule(lr){2-3} \cmidrule(lr){4-5} \cmidrule(lr){6-7} \cmidrule(lr){8-9} \cmidrule(lr){10-11} \cmidrule(lr){12-13} 
& Top1 $\uparrow$ & FID  $\downarrow$ & Top1  $\uparrow$ & FID $\downarrow$ & Top1   $\uparrow$ & FID $\downarrow$ & Top1 $\uparrow$  &  RL $\uparrow$ & Top1   $\uparrow$ & RL $\uparrow$ & Top1  $\uparrow$ & RL $\uparrow$ \\
\midrule
MotionVerse-wo-TMT &75.6 & 0.011 & 53.8 & 0.071 & 70.2 & 0.025 &82.5 &38.2 &45.8 & 25.5 &  67.4 & 39.0  \\
MotionVerse (T5-Base) & 78.5 & \textbf{0.010} & \textbf{56.3} & \textbf{0.064} & 71.9 & \textbf{0.023} & \textbf{84.2} & \textbf{38.8} & \textbf{49.9} & \textbf{25.9} & \textbf{70.2} & \textbf{39.1} \\
\midrule
MotionVerse (T5-Small) &69.1 &0.016 &45.3 & 0.065 & 60.3 & 0.032 & 81.3 & 38.2 &48.6 & 25.2 & 65.1 & 24.7 \\
MotionVerse (T5-Large) &\textbf{79.2} &0.011 &52.1 & 0.074 & \textbf{73.1} &0.028 & 83.4 &38.4 & 48.2 & \textbf{25.9} & 65.9 &37.8 \\
\bottomrule
\end{tabular}
\label{tab_A6}
\end{table*}
\begin{table*}[t]
\centering
\caption{Ablation study of the Task-Level Motion Tower (TMT) and Large Language Backbone on Motion Prediction and Motion In-between tasks.}
\begin{tabular}{l c c c c c c c c c c c c  }
\toprule
\multirow{2}{*}{Models} & \multicolumn{6}{c}{Motion Prediction} & \multicolumn{6}{c}{Motion In-between}  \\
\cmidrule(lr){2-7}  \cmidrule(lr){8-13}   
&\multicolumn{2}{c}{MotionX} &\multicolumn{2}{c}{InterHuman} &\multicolumn{2}{c}{InterX} &\multicolumn{2}{c}{MotionX} &\multicolumn{2}{c}{InterHuman} &\multicolumn{2}{c}{InterX}  \\
\cmidrule(lr){2-3} \cmidrule(lr){4-5} \cmidrule(lr){6-7} \cmidrule(lr){8-9} \cmidrule(lr){10-11} \cmidrule(lr){12-13} 
& FID  $\downarrow$ & ADE  $\downarrow$ & FID  $\downarrow$ & ADE  $\downarrow$ & FID  $\downarrow$ & ADE  $\downarrow$ & FID  $\downarrow$ & ADE  $\downarrow$ & FID  $\downarrow$ & ADE  $\downarrow$ & FID  $\downarrow$ & ADE  $\downarrow$  \\
\midrule
MotionVerse-wo-TMT &0.012  &2.137 &0.050 & 1.375 & \textbf{0.036} & 0.918 &\textbf{0.007} &1.781 &0.043 & 1.187 & 0.035 & 0.837 \\
MotionVerse (T5-Base) &\textbf{0.011} & 2.136 & \textbf{0.042} & 1.329 & \textbf{0.036} & \textbf{0.914} & \textbf{0.007} & 1.780 & \textbf{0.037} & 1.160 & \textbf{0.031} & 0.837  \\
\midrule
MotionVerse (T5-Small)  & 0.022 & 2.630 &0.056 &1.434 & 0.075&0.954 & 0.012 & 2.070 & 0.040 & 1.254 & 0.054 & 0.864 \\
MotionVerse (T5-Large) &0.012 &\textbf{2.119} & 0.051 & \textbf{1.286} & \textbf{0.036} & 0.918 &0.008 &\textbf{1.762} & 0.045 & \textbf{1.146} & \textbf{0.031} & \textbf{0.824}\\
\bottomrule
\end{tabular}
\label{tab_A7}
\end{table*}
\begin{table}[t]
\centering
\caption{Ablation study of the Task-Level Motion Tower (TMT) and Large Language Backbone on Motion Reaction and Editing tasks. MV denotes MotionVerse. }
\begin{tabular}{l c c c c c c  }
\toprule
\multirow{2}{*}{Models} & \multicolumn{4}{c}{Motion Reaction} & \multicolumn{2}{c}{Motion Editing}  \\
\cmidrule(lr){2-5}  \cmidrule(lr){6-7}   
 &\multicolumn{2}{c}{InterHuman} &\multicolumn{2}{c}{InterX} &\multicolumn{2}{c}{MotionFix}  \\
\cmidrule(lr){2-3} \cmidrule(lr){4-5} \cmidrule(lr){6-7} 
& Top1 $\uparrow$ & FID  $\downarrow$    & Top1   & FID  & Top1   & FID \\
\midrule
MV-wo-TMT &50.0 & 0.043 & 70.9 & 0.018 & 37.9 & 0.054 \\
MV (T5-Base) &50.3 & 0.042 & \textbf{71.6} & \textbf{0.017} & 38.7 & \textbf{0.029}  \\
\midrule
MV (T5-Small) & 49.7 & 0.044 & 63.0 & 0.024 & 38.0 & 0.032 \\
MV (T5-Large) &\textbf{50.7} &\textbf{0.040} &69.7 &\textbf{0.017} & \textbf{41.2}&0.031 \\
\bottomrule
\end{tabular}
\label{tab_A8}
\end{table}

\vspace{0.5\baselineskip} 
\noindent
\textbf{The effect of Delay Parallel Modeling.} \
In this part, we assess the effectiveness of different multi-stream motion tokens modeling strategies, namely Flattening, Parallel, and Delay Parallel. %as introduced in Sec. \ref{sec:3.2}.
The evaluation is conducted under the single-task Text-to-Motion (T2M) setting on the MotionX dataset. As shown in Tab. \ref{tab_A2}, the \textit{Parallel} pattern, which predicts all motion codes from the same time step in parallel, performs poorly in motion generation. This is likely due to its inability to capture the hierarchical residual dependencies across quantization levels. In contrast, both \textit{Flattening} and \textit{Delay Parallel} achieve significantly better generation performance. However, the \textit{Flattening} incurs substantial computational overhead, and similar performance can be achieved with the more efficient \textit{Delayed Parallel} strategy. Therefore, we adopt the  \textit{Delayed Parallel} strategy in our approach.

\vspace{0.5\baselineskip} 
\noindent
\textbf{The effect of the Motion-Fusion Module.}
Tab. \ref{tab_A4} evaluates three architecture variants of Motion-Fusion module: Low-RanK Adaptation (LoRA), Mixture-of-Expert (MoE), and Modality-Isolated Architecture (MIS). All variants are trained in a multi-task setting and evaluated on the Text-to-Motion (T2M) task. As shown in Tab. \ref{tab_A4}, the Prototype model exhibits a substantial performance drop in motion generation compared to its single-task training counterpart. We attribute this degradation to modality interference: In the single-task setting, the encoder and decoder of the T5-based language model process text and motion independently, thereby avoiding cross-modal interference. In contrast, under the multi-task setting, both encoder and decoder must jointly handle text and motion modalities to support diverse tasks, leading to modality entanglement and degraded task-specific performance. 

In comparison, our proposed variants adopt modality-specific towers that separately encode motion and text inputs. This architecture decoupling reduces modality interference and facilitates more robust multi-task training. 
As evidenced in Tab. \ref{tab_A4}, on the text-to-motion task, all three variants achieve performance compared to the single-task baseline and consistently outperform the Prototype model (\eg, over $6\%$ gains in top1 retrieval accuracy on InterX.)
These consistent gains validate the effectiveness of the Motion-Fusion module in mitigating modality interference and enhancing task-specific performance.

\vspace{0.5\baselineskip} 
\noindent
\textbf{Comparisons of Different Motion-Fusion Variants.} \
As shown in Tab. \ref{tab_A4}, both MoE and MIS outperform LoRA. This improvement can be attributed to the architectural differences in handle modality interactions. In LoRA, the Motion LoRA and Text LoRA components still share the same weight matrices ($W$ in Eq. \ref{eq15}), which still leads to negative interference between motion and text modalities during multi-task optimization. In contrast, MoE and MIS adopt either independent feedforward networks (FFNs) or fully decoupled parameterization, enabling more effective separation of the two modalities and thereby reducing cross-modal interference. Notably, MoE and MIS exhibit comparable performance, indicating that decoupling FFN alone is sufficient to substantially mitigate modality interference. Considering memory efficiency, we adopt the MoE structure in our implementation. 

\vspace{0.5\baselineskip} 
\noindent
\textbf{The effect of Task-Specific Motion Tower in Multitask Instruction Fine-Tuning.} \ In Tab. \ref{tab_A6}, \ref{tab_A7}, \ref{tab_A8}, we compare MotionVerse and MotionVerse-wo-TMT which excludes the task-specific motion tower during the multitask instruction fine-tuning stage. As shown in these tables, MotionVerse consistently outperforms MotionVerse-to-TMT across all evaluated tasks, demonstrating the effectiveness of the task-specific motion tower in alleviating interference among heterogeneous tasks. Specifically, the motion-to-text task primarily relies on high-level, global motion features, while the motion-to-motion task demands fine-grained modeling of low-level temporal dynamics. Thus introducing a dedicated motion tower for the motion-to-motion task enables the model to better  disentangle task-specific representations, leading to improved performance in the multi-task training setting.

\vspace{0.5\baselineskip} 
\noindent
\textbf{Comparisons of Different Large Language Backbone.} \
In this part, we evaluate the performance of MotionVerse using different T5-based  large language model backbones: T5-small ($\sim60$M parameters), T5-base ($\sim 220$M parameters), T5-large ($\sim 770$M parameters). 
As shown in Tab. \ref{tab_A6}, \ref{tab_A7}, \ref{tab_A8}, T5-Base and T5-Large exhibit comparable overall performance. We attribute this performance parity to a potential saturation effect, wherein the  increased parameter count of T5-Large yields minimal gains for the target tasks. Conversely, the relatively poor performance of T5-Small stems from its constrained model capacity, which proves insufficient for capturing the semantic complexity and task-specific patterns required for our multi-task learning framework. Based on these observations, we adopt T5-base as the backbone language model for MotionVerse, balancing representational power with computational efficiency.

\section{Conclusion}
In this work, we present MotionVerse, a unified framework that harnesses autoregressive LLMs to jointly address motion comprehension, generation, and editing. 
To represent continuous human motion, we propose a residual quantization-based motion tokenizer in conjunction with a Delay Parallel Modeling strategy, which discretizes motion into multi-stream token sequences and enables efficient modeling of inter-stream dependencies within the LLM backbone. 
To mitigate modality interference between motion and language, we adopt a disentangled design that incorporates modality-specific weights within the shared LLM backbone. Extensive evaluations across diverse motion-relevant tasks demonstrate the effectiveness of MotionVerse in motion comprehension, generation, and editing.  
% \appendices
% \section{}

% use section* for acknowledgment
\ifCLASSOPTIONcompsoc
  % The Computer Society usually uses the plural form
  \section*{Acknowledgments}
\else
  % regular IEEE prefers the singular form
  \section*{Acknowledgment}
\fi

This work is partially supported by National Natural Science Foundation of China (NSFC): 62376259 and 62306301.

% Can use something like this to put references on a page
% by themselves when using endfloat and the captionsoff option.
\ifCLASSOPTIONcaptionsoff
  \newpage
\fi

\bibliographystyle{plain}
\bibliography{example_paper}

\begin{IEEEbiography}[{\includegraphics[width=1in,height=1.25in,clip,keepaspectratio]{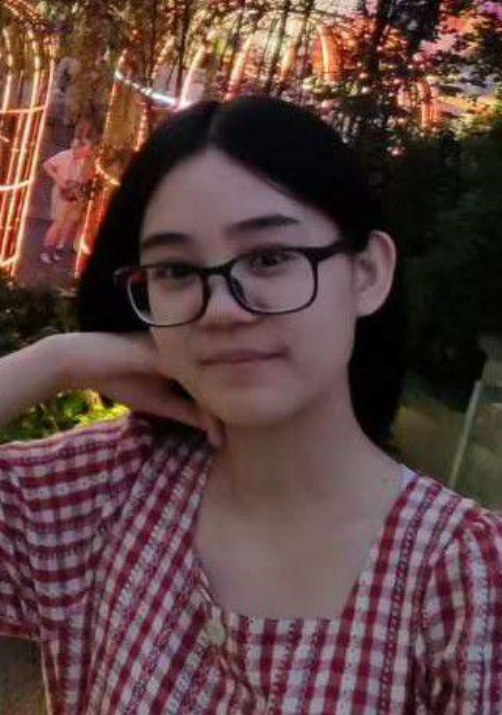}}]{Ruibing Hou} (Member, IEEE) 
received the BS degree in Northwestern Polytechnical University, Xi’an, China, in 2016. She received PhD degree in computer science from the Institute of Computing Technology, Chinese Academy of Sciences, Beijing, China, in 2022. She is currently a Associate Professor with the Institute of Computing Technology, Chinese Academy of Sciences. Her research interests are in machine learning and computer vision. She especially focuses on multimodal large language model and 3D humam modeling. 
\end{IEEEbiography}

\begin{IEEEbiography}[{\includegraphics[width=1in,height=1.25in,clip,keepaspectratio]{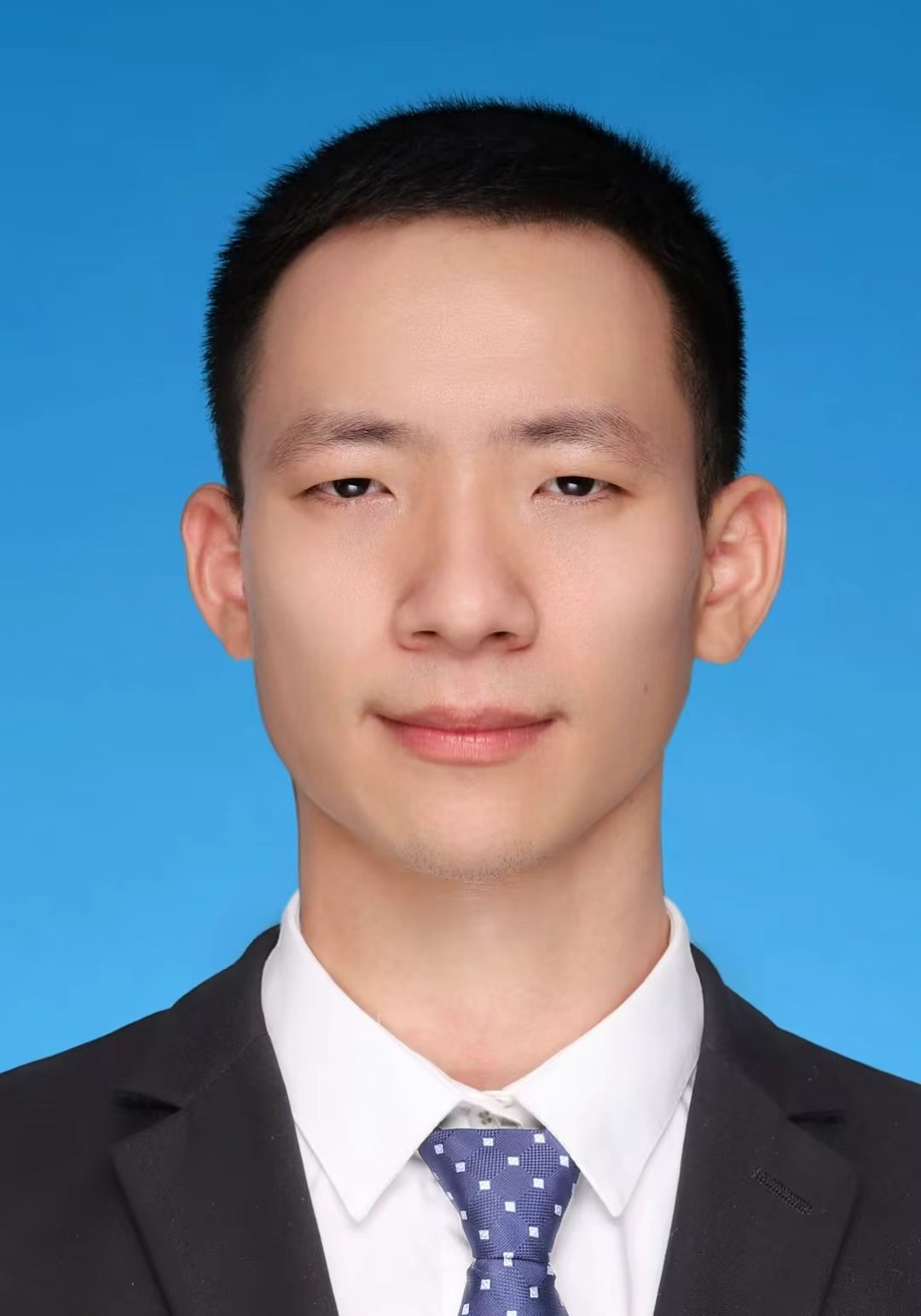}}]{Mingshuang Luo} 
received the BS degree in electric information science and technology from Xiangtan University in 2018; the MS degree in computer technology from the University of the Chinese Academy of Sciences (UCAS) in 2021. He has been pursuing the PhD degree at the Institute of Computing Technology (ICT), Chinese Academy of Sciences (CAS), since 2022. From 2021 to 2022, he worked at Xiaomi Group as a Speech Algorithm Researcher. His research interests are in computer vision, pattern recognition, and machine learning. He especially focuses on human motion generation, human animation generation and related research topics.
\end{IEEEbiography}

\begin{IEEEbiography}[{\includegraphics[width=1in,height=1.25in,clip,keepaspectratio]{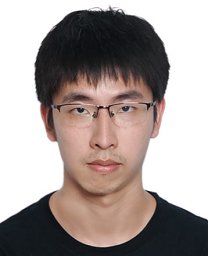}}]{Hongyu Pan}
received the B.E. degree from Beijing Institute of Technology (BIT) in 2016 and
the M.S. degree in computer science from the
Institute of Computing Technology (ICT), University of Chinese Academy of Sciences (UCAS),
in 2019. He is currently an employee at Horizon
Robotics. His research interests include computer vision, pattern recognition, and image processing. He specifically focuses on 3D detection/segmentation/motion and depth estimation.
\end{IEEEbiography}

\begin{IEEEbiography}[{\includegraphics[width=1in,height=1.25in,clip,keepaspectratio]{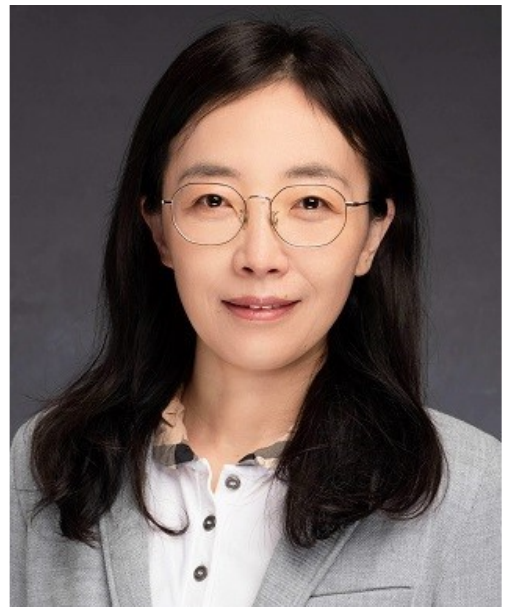}}]{Hong Chang} (Member, IEEE) 
received the bachelor’s degree in computer science from the Hebei University of Technology, Tianjin, China, in 1998, the MS degree in computer science from Tianjin University, Tianjin, in 2001, and the PhD degree in computer science from the Hong Kong University of Science and Technology, Kowloon, Hong Kong, in 2006. She was a research scientist with Xerox Research Centre Europe. She is currently a researcher with the Institute of Computing Technology, Chinese Academy of Sciences, Beijing, China. Her main research interests include algorithms and models in machine learning, and their applications in pattern recognition, computer vision and AI2Science.
\end{IEEEbiography}

\begin{IEEEbiography}[{\includegraphics[width=1in,height=1.25in,clip,keepaspectratio]{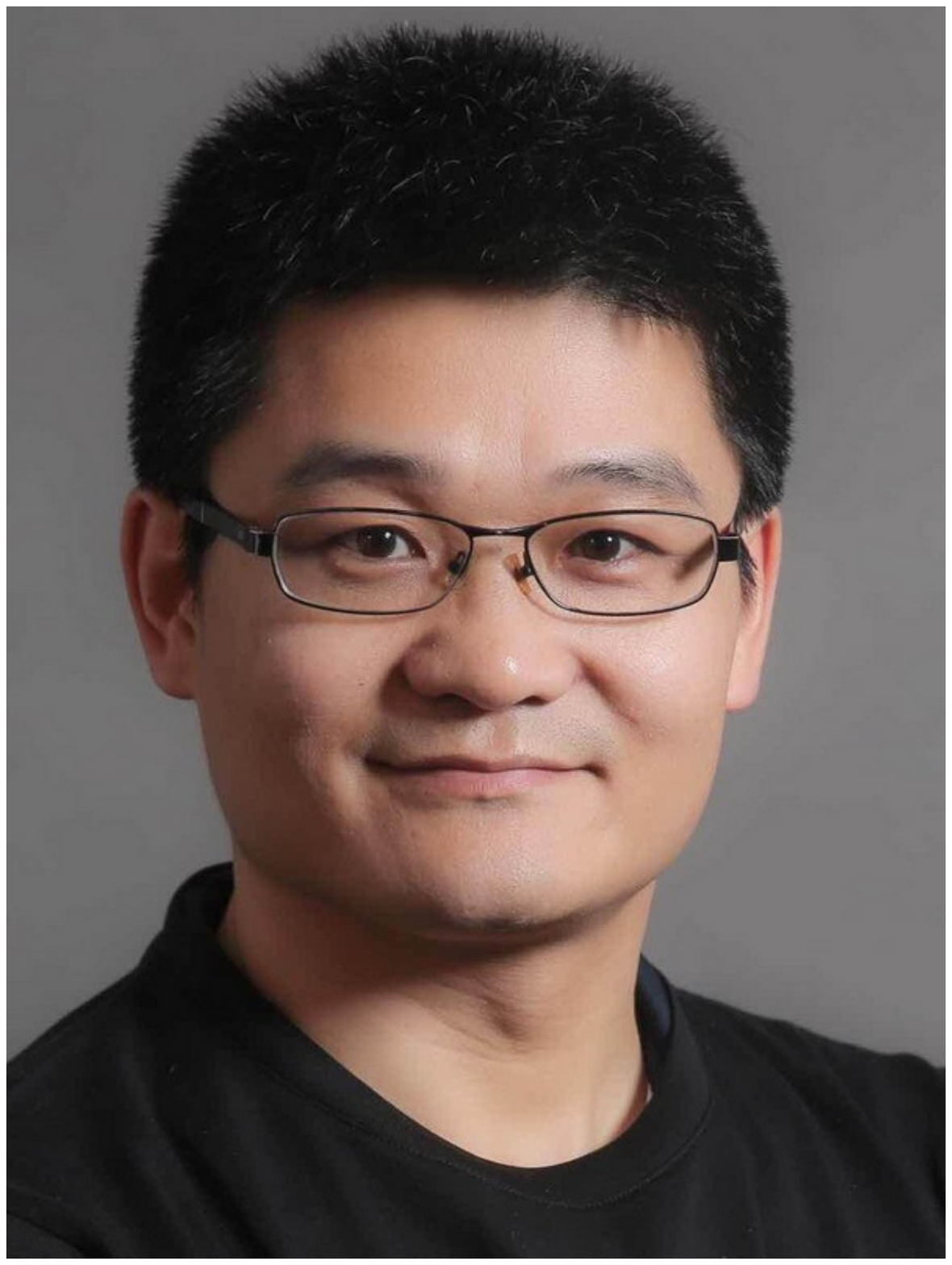}}]{Shiguang Shan} (Fellow, IEEE)
received the PhD degree in computer science from the Institute of Computing Technology, Chinese Academy of Sciences (CAS), Beijing, China, in 2004. Since 2010, he has been a full professor with the Institute of Computing Technology. He is currently the deputy director with the CAS Key Lab of Intelligent Information Processing. He has authored or coauthored more than 300 papers, with totally more than 20,000 Google scholar citations. His research interests include computer vision, pattern recognition, and machine learning. He was an area chairs for many international conferences including CVPR, ICCV, AAAI, IJCAI, ACCV, ICPR, and FG. He is/was associate editors of several journals including IEEE TRANSACTIONS ON IMAGE PROCESSING, Neurocomputing, Computer Vision and Image Understanding, and PRL. He was the recipient of the China’s State Natural Science Award in 2015 and China’s State S\&T Progress Award in 2005 for his research work.
\end{IEEEbiography}

% that's all folks
\end{document}